\documentclass[runningheads]{llncs}
\usepackage[T1]{fontenc}
\usepackage{graphicx}
\usepackage{booktabs}
\usepackage{microtype}
\usepackage[misc]{ifsym}
\newcommand{\corr}{(\Letter)}
\usepackage{mwe}
\usepackage{amsmath}
\usepackage{graphicx}
\usepackage[colorlinks=true, allcolors=blue]{hyperref}
\usepackage{algorithm}
\usepackage{algpseudocodex}
\usepackage{amssymb}

\usepackage{booktabs}
\usepackage{multirow}
\usepackage{graphicx}
\usepackage{framed}

\algnewcommand\algorithmicforeach{\textbf{for each}}
\algdef{S}[FOR]{ForEach}[1]{\algorithmicforeach\ #1\ \algorithmicdo}

\begin{document}

\title{A robust methodology for long-term \\ sustainability evaluation of Machine Learning models}

\titlerunning{Robust methods for evaluating long-term ML sustainability}

\author{Jorge Paz-Ruza\inst{1}\corr \and
João Gama\inst{2} \and
Amparo Alonso-Betanzos\inst{1} \and Bertha Guijarro-Berdiñas\inst{1}}

\authorrunning{Paz-Ruza J. et al.}

\institute{Universidade da Coruña, LIDIA - CITIC, Campus de Elviña s/n, 15071 A Coruña, Spain \email{\{j.ruza, amparo.alonso.betanzos, berta.guijarro\}@udc.gal}
\and
Faculty of Economics, University of Porto, Porto, 4200-465, Portugal \email{jgama@fep.up.pt}}

\maketitle              

\begin{abstract}
    Sustainability and efficiency have become essential considerations in the development and deployment of Artificial Intelligence systems, but existing regulatory practices for Green AI still lack standardized, model-agnostic evaluation protocols. Recently, sustainability auditing pipelines for ML and usual practices by researchers show three main pitfalls: 1) they disproportionally emphasize epoch/batch learning settings, 2) they do not formally model the long-term sustainability cost of adapting and re-training models, and 3) they effectively measure the sustainability of sterile experiments, instead of estimating the environmental impact of real-world, long-term AI lifecycles. In this work, we propose a novel evaluation protocol for assessing the long-term sustainability of ML models, based on concepts inspired by Online ML, which measures sustainability and performance through incremental/continual model retraining parallel to real-world data acquisition. Through experimentation on diverse ML tasks using a range of model types, we demonstrate that traditional static train-test evaluations do not reliably capture sustainability under evolving datasets, as they overestimate, underestimate and/or erratically estimate the actual cost of maintaining and updating ML models. Our proposed sustainability evaluation pipeline also draws initial evidence that, in real-world, long-term ML life-cycles, higher environmental costs occasionally yield little to no performance benefits.

\end{abstract}

\section{Introduction}

One of the main effects of the increasing use of Artificial Intelligence (AI) and Machine Learning (ML)-based systems on everyday life is the computational cost and environmental impact they entail given their high-intensity energy use \cite{dingsong2025tracking,oecd2022measuring,ecb2025increasing}; recent research has estimated the current ``power hunger'' of AI models in datacenters accounts for half of their total global demand, equivalent to double the yearly energy use of the Netherlands \cite{guardian2025ai_datacentre_power}.

Among the many desirable properties of AI systems, sustainability and efficiency have become increasingly important; this is not only because of the global context of a warming climate, water scarcity and energy supply difficulties, but also due to the need for simpler, more efficient models in resource-constrained settings, like IoT or local embedded models \cite{schizas2022tinymllowpower}. Both academic and non-academic actors have progressively realized this need, and encouraged researchers, companies and practitioners to promote ML sustainability. Recent scholarly literature has brought the eye into the environmental cost of AI research and development \cite{agarwal2025hidden_ai_race,UNESCO_AI_Recommendation_2021,singh2024survey_sustainability_llms}, and corporate and public initiatives to foster ``Green AI'' have begun to emerge \cite{ai_ei_principles2025,microsoft_responsible_ai_standard2025}.

Perhaps most crucially, there have been attempts on official, government-level regulating the sustainability of AI as a core ``ethical property'': the EU AI Act directs that the sustainability of AI (in terms of its environmental and social footprint) should be considered when developing and deploying AI pipelines \cite{AI_Act}; the proposal of the ``AI Environmental Impacts Act'' in the US would require the study of AI’s environmental impacts, including potential increases in energy consumption from computation, data centers, and infrastructure \cite{us_ai_environmental_impacts_act2024}. 

However, this consensual agreement on the importance of sustainability and efficiency for real-world AI systems and the social and regulatory efforts heavily contrasts with the practical applicability of such regulations; without looking further, the AI Act itself defines sustainability as a need, but does not indicate what measurement protocols, lifecycle frameworks, reporting methodology or carbon accounting standards should be considered \cite{AI_Act}; these should be certainly necessary considerations to ensure a robust, reliable and practical assessment of AI models' environmental impact in actual, long-term operation.

We argue that this lack of comprehensiveness in sustainability recommendations for AI systems does not stem from a careless or sloppy construction of the regulations themselves, which are well-intended and intentionally designed on a high level of abstraction. Instead, this methodological gap arises from an actual absence of suitable evaluation protocols that are formal, model-agnostic, reproducible, and grounded in real-life usage protocols for the ML lifecycle. 

While there exist different regional and international proposals or technical reports for standardisation of AI sustainability assessment, a comprehensive inspection reveals these are in practice focused on proposing what aspects and units of metrics ought to be measured (e.g. Kw/h of energy, gCO$_2$eq of emissions, m$^3$ of employed water on cooling), defining the different scopes of emissions or defining what constitutes a Life Cycle Assessment. The AI Action Summit report, which centralized and reviewed different ISO, ITU and other organizations' standards, directly stated that there still exist \textit{``a number of normative gaps relating to AI sustainability''} and that, beyond mere unit definitions, it is still necessary to ``develop common environmental indicators that are measurable or can be estimated for each lifecycle stage of AI system resources'' \cite{itu_aiforgood_standardization_ai_sustainability2025}.

Although many researchers and companies have made it a habit to report efficiency figures and comparisons (e.g., in terms of emitted $CO_2$), we identify two major pitfalls in how this is done in the literature:

\begin{itemize}
    \item \textbf{Experiment-bound sustainability assessment}: Authors and development teams measure what is effectively the sustainability of the environmental impacts of the experiments, rather than the AI system itself or its lifecycle. This is generally caused by directly measuring efficiency on research, experiment-focused train-evaluation pipelines, such as directly measuring the emitted $CO_2$ in individual train-test holdout experiments or cross-validation procedures.
    \item \textbf{Generalizability of the evaluation protocol}: sustainability is measured in such a way that is not model-agnostic, often biased towards popular neural network and batch-oriented methodologies in content generation (e.g. generative LLMs) or classification tasks; several of the proposed standards mention concepts like ``per-token'' or ``per-epoch'' environmental costs, indicators only applicable to a fraction of the real-world ML tasks and models \cite{une0086}.
\end{itemize} 

In this research, we set the foundations for a fair and reliable evaluation ground for the \textit{long-term} sustainability of models, in a way that is standardised, reliable, model-agnostic and representative of real-world long-term usage of deployed ML pipelines. Our proposal is centered on indicators related to the greenhouse emissions of the models themselves, and is based on several reasonable assumptions: 1) a model's sustainability depends not only on its theoretical complexity, but its ability to adapt to and leverage new data, 2) models perform inference on a superset of the eventual training data, even if development/optimization experiments are ``frozen in time'', and 3) in most real-world tasks, ML models will indeed require retraining or continuous training with this new data, either to obtain the desired performance or to avoid an opportunity cost. 

Our proposed evaluation protocol for ML sustainability is inspired by the \textit{prequential} evaluation of Streaming ML: we suggest employing current available data in increasingly large instance subsets to analyze the \textit{cumulative} environmental impact of performing inference on every new instance while intermittently or periodically retraining or continuously training the model on now labelled instances. This allows practitioners to carefully gauge indicator trade-offs in the \textit{performance vs. environmental impact vs. dataset size} triad on a long-term scale, and anticipate potentially unsustainable ML models on production-level requirements on model efficiency or performance.

We performed computational experiments on four classic ML tasks, using ML models of varying natures and complexities (ensembles, neural networks, pre-trained models), and compared our proposed evaluation pipeline to performing isolated sustainability measurements on traditional holdout or cross-validation experiments. Our findings prove our initial assumptions, in that:

\begin{itemize}
    \item The evaluation of a static training and test protocol, even if further inference into the future is considered, is not representative of the actual long-term sustainability of a Machine Learning model, either underestimating, overestimating and/or erratically measuring CO$_2$ and its relation with model complexity, performance or task and data characteristics.
    \item Beyond performance and short-term sustainability, long-term sustainability of models should play a huge role in the decision-making of selecting the most adequate ML model for an application, as some models exhibit long-term exponential increases in environmental impact with marginal or null gains in performance. 
\end{itemize}
\section{Existing Standards and Protocols for Measuring ML Sustainability}

Existing initiatives to standardise and promote the evaluation of ML models' sustainability differ substantially in scope, granularity, and intended use. This section reviews existing related standards and discusses limitations of those which at present propose low-level methodologies for measuring environmental impact.

We choose to distinguish, based on their nature, four types of standards, protocols or metrics in the existing ecosystem of Sustainable ML regulation:

\begin{itemize}

\item \textbf{Energy and Carbon Reporting in ML Research}:  
Most academic works, either in position or in practice, advocate for reporting experimental energy, runtime, and CO$_2$ emissions ~\cite{strubell2019energy,henderson2020systematic}, but fail in capturing long-term and holistic comparisons.

\item \textbf{Carbon Tracking Tools (e.g., CarbonTracker, CodeCarbon)}~\cite{lacoste2020carbontracker}   
provide reliable and model-agnostic measurements of CO$_2$-equivalent emissions, but still depend on local instrumentation, leaving the actual experimental design to the practitioner or the researcher.

\item \textbf{AI Lifecycle Standards (e.g., ISO/IEC 5338, ISO/IEC TR 20226, etc.)}~\cite{iso5338,isoiec20226,afnor2314,cen18145} which define AI system lifecycle processes and discuss environmental sustainability aspects. Their scope is governance-oriented and high-level, primarily addressing organizational or service-level assessment rather than standardized model-level experimental evaluation.

\item \textbf{General sustainability standards applied to ML}, such as the GHG Protocol~\cite{ghgprotocol} defining Scope~1–3 greenhouse gas accounting principles, ISO 14040/14044 Life Cycle Assessment standards~\cite{iso14040,iso14044}, or Data Center Efficiency Metrics (e.g., EN 50600)~\cite{en50600}. These frameworks provide methodological foundations for emissions accounting and lifecycle reasoning, yet they lack model-level granularity and standardized functional units aligned with ML benchmarking practices.

\end{itemize}

Among AI and ML-specific standards, the very recent UNE 0086:2025 specification~\cite{une0086} sought to be the first formalized, certifiable standards dedicated to environmental impact assessment of AI systems. It defines explicit system boundaries, metrics and reporting requirements across three lifecycle components with respect to carbon emissions:

\begin{enumerate}
    \item \textbf{Operational energy consumption} during development, training, and inference.
    \item \textbf{Carbon conversion} using geographically appropriate emission factors.
    \item \textbf{Embodied emissions allocation} from hardware manufacturing and infrastructure, proportionally assigned to the AI system.
\end{enumerate}

Despite its rigour, there exist several limitations that affect generalizability and its applicability as an industry and institutional-level norm in the broader AI ecosystem. We identify the most relevant ones as:

\begin{itemize}
\item It assumes strictly separable and time-disjoint training and inference phases, which is problematic for any continual or adaptive learning system.
\item There is a clear orientation towards massive Large Language Model assessment, given the mandatory reporting of metrics as ``per-batch emissions'' and ``per-epoch emissions'', ``energy per model parameter'', etc. as well as the focus on pretraining, finetuning, and foundational model costs, which do not apply to much of the ML landscape.
\item Life-cycle-level assessments are still defined in purely descriptive terms, once again delegating the long-term sustainability analysis modelling and experimentation to the researcher or practitioner. Also, it depends on hardware lifetime and utilization assumptions, introducing variability across deployments.
\end{itemize}

Thus, while UNE 0086:2025 first provides an auditable carbon accounting structure, it does not define a sufficiently formal methodology, lacking formality in defining long-term sustainability assessment, and how to effectively analyze performance-sustainability trade-offs. Coincidentally, the officially released emissions calculator for the UNE 0086:2025 standard \cite{calculadorapnav2025} in practice wraps \textit{CodeCarbon} to segregate the emissions into training and inference and compute the required metrics; the practitioner is still fully responsible for measuring long-term ML sustainability in a way that does not only represent the current sterile experiment's emissions. 

This gap motivates the need for a harmonized train-and-evaluation protocol capable of jointly assessing long-term model performance and environmental impact, which we tackled with our proposed method. 
\section{Proposed Protocol}

The goal of this research is to design a reliable, robust evaluation protocol for assessing long-term sustainability that is model-agnostic and is representative of the actual environmental lifecycle of an AI system. Instead of assuming the nature of characteristics of AI models, as done by UNE 0086:2025, we generalize the task to evaluating the efficiency of a model $\phi: x \rightarrow y$ which has, fundamentally, two available operations:

\begin{itemize}
    \item $train(\phi,D)\rightarrow \phi$, which trains $\phi$ on an instance or set of instances $D$
    \item $\phi(x) \rightarrow y$, which performs inference on data point $x$ employing $\phi$
\end{itemize}

\noindent and to respect the likely real-time use-case of real-world ML systems, where immediate inference is required, we assume that instances arrive incrementally rather than in batches, which is crucial for proper efficiency evaluation. By using this high level of abstraction we avoid focusing on per-batch or per-epoch costs, which is directly causal to long-term efficiency and cannot be applied to a majority of learning paradigms; we also avoid blindly using static per-inference costs to assess inference sustainability, as the computational cost of performing $\phi(x)$ can vary with the evolution of $\phi$ (e.g. increasingly deep decision trees).

We propose to consider the fact that, regardless of the nature of the models and the data, models will need periodic or incremental retraining or updating to keep up with evolving data, achieve a desired level of performance, or avoid opportunity costs of operation or against competitors; simultaneously, models must still perform inference of all incoming examples. Given the available data, we incrementally present the models with all instances for inference, and consider that the conditions and/or intervals for model retraining and evaluation will be dependent on context (model and data) to be identified at the beginning of the ML's developmental lifecycle. For our initial experiments, we make a reasonable assumption that models should be evaluated each time the number of instances increases by 50\%. Batch models are retrained on the same basis, while Streaming models can be updated on every example.

Algorithms \ref{alg:sus-eval-batch} and \ref{alg:sus-eval-streaming} show our proposed evaluation protocol for the long-term assessment of sustainability and performance trade-offs, making sure it is compatible with both offline (batch) and online (streaming) learning approaches, which is typically not possible for state-of-the-art evaluation methodologies.

\begin{algorithm}[h]
\scriptsize
\caption{Long-term evaluation (Batch)}
\begin{algorithmic}[5]

\Require
\Statex $S$: Continuous data stream
\Statex $\phi$: ML model
\Statex $n$: Initial no. of examples
\Statex $w$: Size of evaluation window
\Statex $\lambda$: Re-training frequency growth factor
\Ensure
\Statex Cumulative evaluation results $R \in \{(t_0, e_0, \mathcal{P}_0), (t_1, e_1, \mathcal{P}_1), ...)\}$ s.t. $t_k = n\lambda^k$ 

\item[]
\State Initialize $T \gets \emptyset$
\State Initialize evaluation window $W$ with size $w$
\ForAll{$s = (x, y) \in S$}
\State $\hat{y} \gets \phi(x)$
\State Add $(y, \hat{y})$ to W
\State Add $s$ to $T$
\If{$|T|$ = $n$}
\State $\phi \gets train(\phi,T)$
\State $\mathcal{P} \gets$ performance metrics over $W$
\State $e \gets$ sustainability metrics (accumulated gC0$_2$e)
\State \textbf{yield} $n, e, \mathcal{P}$
\State $n \gets \lambda n$
\EndIf
\EndFor
\end{algorithmic}
\label{alg:sus-eval-batch}
\end{algorithm}

\begin{algorithm}[h]
\scriptsize
\caption{Long-term evaluation (Streaming)}
\begin{algorithmic}[5]

\Require
\Statex $S$: Continuous data stream
\Statex $\phi$: ML model
\Statex $n$: Initial no. of examples
\Statex $w$: Size of evaluation window
\Statex $\lambda$: Evaluation frequency growth factor
\Ensure
\Statex Cumulative evaluation results $R \in \{(t_0, e_0, \mathcal{P}_0), (t_1, e_1, \mathcal{P}_1), ...)\}$ s.t. $t_k = n\lambda^k$ 

\item[]
\State Initialize evaluation window $W$ with size $w$
\ForAll{instance $s=(x, y) \in S$}
\State $\hat{y} \gets \phi(x)$
\State Add $(y, \hat{y})$ to W
\State $\phi \gets train(\phi,s)$
\If{$i = n$}
\State $\mathcal{P} \gets$ performance metrics over $W$
\State $e \gets$ sustainability metrics (accumulated gC0$_2$e)
\State \textbf{yield} $n, e, \mathcal{P}$
\State $n \gets \lambda n$
\EndIf
\EndFor
\end{algorithmic}
\label{alg:sus-eval-streaming}
\end{algorithm}

By definition, this evaluation approach provides a fair ground by allowing any number of models to be tested in the same data while still providing flexibility for the intricacies and advantages of each ML model or paradigm (such as the continual prequential learning of streaming systems). The resulting metrics yielded sequentially by the system ($R \in \{(t_0, e_0, \mathcal{P}_0), (t_1, e_1, \mathcal{P}_1), ...)\}$ s.t. $t_k = n\lambda^k$) will most practically define curve trade-offs (number of examples vs performance vs. emissions), with important significance for the life cycle assessment of the model:

\begin{itemize}
    \item An \textbf{no. of instances vs. emissions} trade-off is indicative of the practical feasibility of models on production-level scales of training data, like those that require billions of samples for training effectively (e.g., LLMs).
    \item A \textbf{no. of instances vs performance} trade-off can guide the data requirements and estimate potential environmental costs of data gathering in prior and posterior phases of the ML lifecycle.
    \item An \textbf{emissions vs. performance} trade-off can help identify more sustainable choices in energy-conscious settings or environments with limited operational resources, as well as help avoid massive model carbon footprints if only marginal gains are being achieved.
\end{itemize}
\section{Experimental setup}

This section provides basic information about the datasets, models and other implementation details used to evaluate our methodology for long-term ML sustainability assessment. 

\subsection{Datasets and models}
To prove the suitability of our sustainability evaluation approach to different domains, we select 9 different datasets across four classic ML tasks (classification, regression, recommendation and anomaly detection) of different scales, feature types and feature dimensionality; nevertheless, we are confident that this approach should be readily applicable to any arbitrary ML task;  basic information about these datasets is shown in Table \ref{tab:datasets}.

Regarding models, Table \ref{tab:datasets} also indicates the model types employed for each dataset. Note that we have selected models that either have directly comparable and equivalent batch and streaming architectures (like tree ensembles), or can be used in streaming or batch form interchangeably (like neural network-based models). In particular, we test, for each selected dataset and model type:

\begin{itemize}
    \item Our long-term sustainability evaluation in the model's Streaming learning form (see Algorithm \ref{alg:sus-eval-streaming}), as a grounded reference for a real-life incrementally learning and cumulatively costly algorithm. 
    \item An ``uninformed'' non-cumulative measurement of retraining the model at the different reevaluation/retraining steps, considering 1) train-test holdout and 2) a 10-fold cross-validation. This represents the simplest and currently most common and popular way of measuring sustainability, i.e., measuring the environmental impact of the experiment rather than the model's long-term life-cycle.
    \item Our long-term sustainability evaluation in the model's Batch form (see Algorithm \ref{alg:sus-eval-batch}), as our proposal for a more suitable and representative long-term evaluation of a model's environmental impact. 
\end{itemize}

\begin{table}[]
\caption{Relation of datasets and models employed to assess the proposed ML sustainability evaluation pipeline on different classic ML tasks (classification, regression, recommendation and anomaly detection).}
\label{tab:datasets}
\resizebox{\textwidth}{!}{%
\begin{tabular}{lllrrll}
\hline
\textbf{Task}                                                & \textbf{Dataset}    & \textbf{Nature} & \multicolumn{1}{l}{\textbf{Instances}} & \multicolumn{1}{l}{\textbf{Features}}                        & \textbf{Target} & \textbf{Models}                                                                              \\ \hline
\multirow{4}{*}{Classification}                              & Waveform-40 ND      & Synthetic       & 1,000,000                              & 40                                                           & Multi-class     & \begin{tabular}[c]{@{}l@{}}Bagging ensemble\\ Boosting ensemble\\ MLP\end{tabular}           \\ \cline{7-7} 
                                                             & QMNIST              & Real            & 402,953                                & 784                                                          & Multi-class     & \begin{tabular}[c]{@{}l@{}}CNN\\ Pretr. Resnet + MLP\end{tabular}                            \\ \cline{7-7} 
                                                             & NSL-KDDCUP99        & Real            & 148,517                                & 41                                                           & Binary          & \begin{tabular}[c]{@{}l@{}}Bagging ensemble\\ Boosting ensemble\\ MLP\end{tabular}           \\ \cline{7-7} 
                                                             & Movielens 1M        & Real            & 1,000,029                              & 42                                                           & Multi-class     & \begin{tabular}[c]{@{}l@{}}Bagging ensemble\\ MLP\end{tabular}                               \\ \hline
\multirow{2}{*}{Regression}                                  & Year Prediction MSD & Real            & 515,345                                & 90                                                           & Real            & \begin{tabular}[c]{@{}l@{}}Bagging ensemble\\ Boosting ensemble\\ MLP\end{tabular}           \\ \cline{7-7} 
                                                             & Automotive          & Real            & 1,000,000                              & 19                                                           & Real            & \begin{tabular}[c]{@{}l@{}}Bagging ensemble\\ Boosting ensemble\\ MLP\end{tabular}           \\ \hline
\multirow{2}{*}{Recommendation}                              & Movielens 1M        & Real            & 575,280                                & \begin{tabular}[c]{@{}r@{}}2(IDs) \\ 42(feat.)\end{tabular} & Ranking         & \begin{tabular}[c]{@{}l@{}}Item-Item (IDs)\\ MF (IDs)\\ NCF (IDs)\\ MLP (feat.)\end{tabular} \\ \cline{7-7} 
                                                             & Netflix             & Real            & 100,000,000                            & 2(IDs)                                                      & Ranking         & \begin{tabular}[c]{@{}l@{}}Item-Item (IDs)\\ MF (IDs)\\ NCF (IDs)\end{tabular}               \\ \hline
\begin{tabular}[c]{@{}l@{}}Anomaly \\ Detection\end{tabular} & CelebA              & Real            & 202,598                                & 39                                                           & Normal/Anomaly  & \begin{tabular}[c]{@{}l@{}}Isolation Forest\\ Autoencoders \\ Nearest Neighbour\end{tabular}                      \\ \hline
\end{tabular}%
}
\end{table}

\subsection{Implementation details}

In the following, we list some fine-grained implementation details relevant for experiment reproducibility and to understand the robustness of the experimental findings detailed in the next section: 

\begin{itemize}
    
    \item Ensemble models all use Java backend implementations (\textit{Weka} for batch, \textit{MOA} for streaming). Neural network models all use PyTorch implementations with no use of GPU, except for the ResNet-MLP pipeline on QMNIST (due to the unaffordable cost of running the batch algorithm on CPU alone). 
    
    \item All MLPs feature two hidden layers with 32 and 16 neurons, and inter-layer ReLu activations. CNNs use two convolution layers of sizes 8x8 and 4x4, with inter-layer batch normalization and max-pooling. All NN's employ learning rate $10^{-3}$ and batch approaches utilize batch sizes of $2^{10}$ and early stopping with patience $p=3$ and $\Delta=10^{-4}$. 
    \item All non-NN models employ the default hyperparameters of their respective libraries (ensuring that, for any given model, common hyperparameters are set identically for their batch and streaming equivalent algorithms). Given the number of datasets and models considered, we refer readers to the official GitHub repository of the project for a comprehensive inspection of model hyperparameters.\footnotemark{}
    \item All experiments were run on a dedicated machine with Windows 10 OS, 32GB RAM, NVIDIA 2060 RTX GPU and an Intel i7-10700 CPU. 

    \item Initial hyperparameters of Algorithms \ref{alg:sus-eval-batch} and \ref{alg:sus-eval-streaming} were set to $\lambda=1.5$, $n=100$, $w=10000$.
    \item Measurements of $CO_2$ emissions were taken employing the \textit{CodeCarbon} library with all logging disabled and modifications to minimize the computational cost of executing the measurements, and fixed Carbon Intensity (100\%) for all the experiments. 
\end{itemize}

\footnotetext{https://github.com/Kominaru/eclair}

Our experimental pipeline is published in open source as a GitHub repository\footnotemark[\value{footnote}], which also includes our proposed ML sustainability evaluation protocol as a library-independent, ready-to-use Python tool. 
\section{Results}

\begin{figure}[!htbp]
    \centering
    \textbf{Waveform 40 \\}
    \includegraphics[width=.30\linewidth]{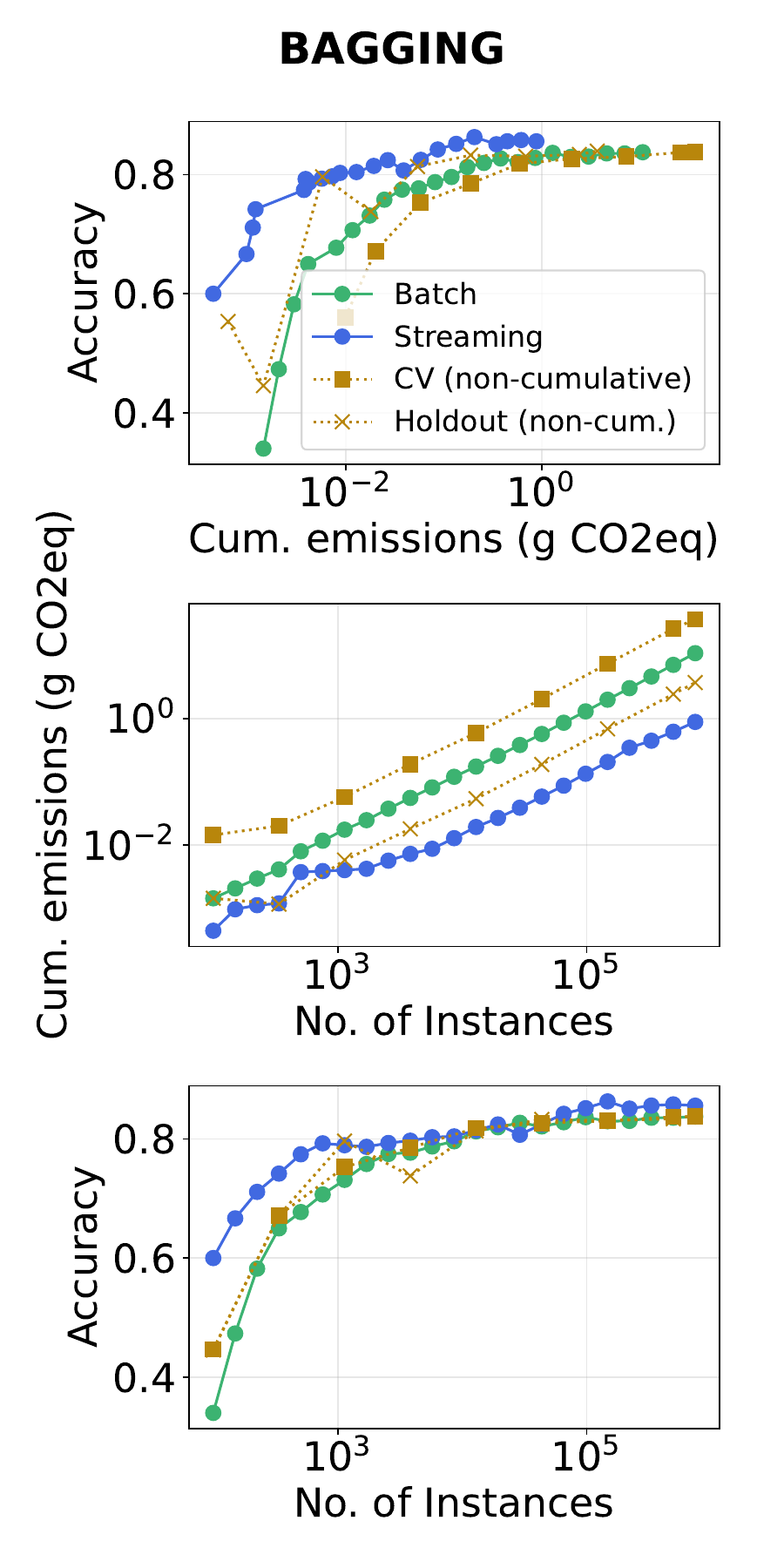}
    \includegraphics[width=.30\linewidth]{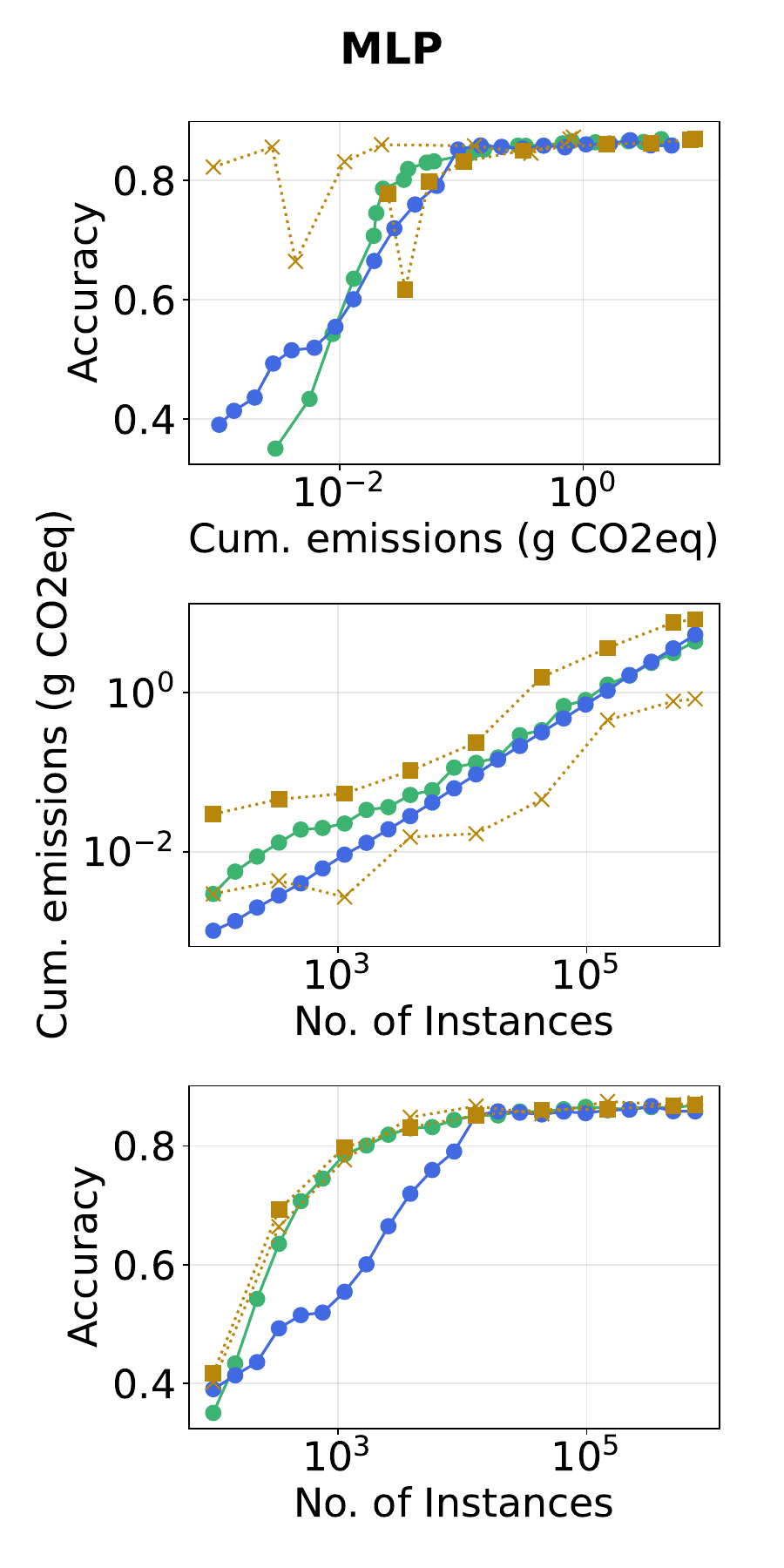}
    \includegraphics[width=.30\linewidth]{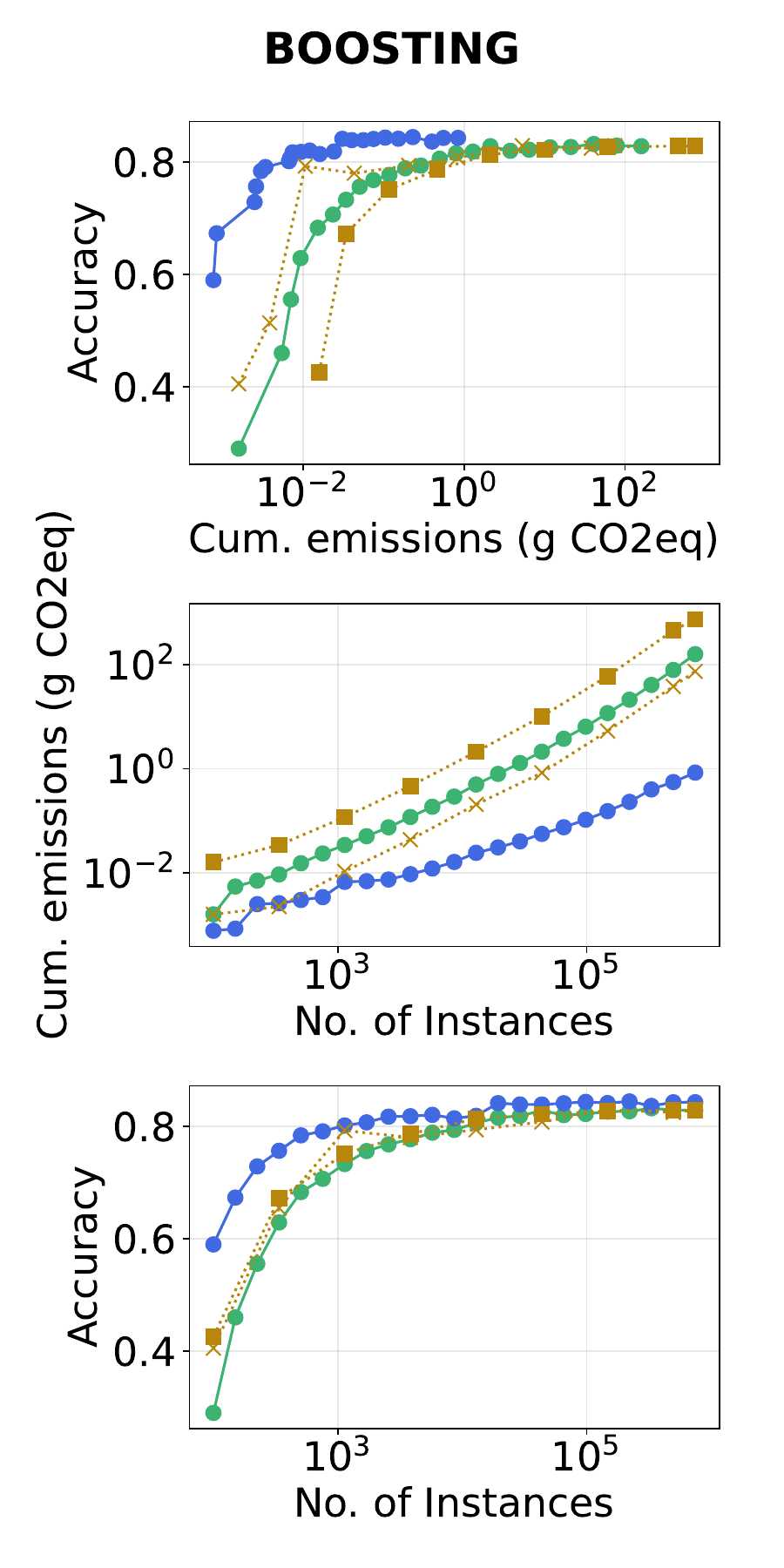}
    \caption{Sustainability vs. No. of instances vs. Performance for our proposal and existing naive experimental approaches on the Waveform 40 dataset.}
    \label{fig:results_waveform}
\end{figure}

\begin{figure}[!h]
\centering

\begin{minipage}[c]{0.48\linewidth}
    \centering
    \textbf{QMNIST}\\
    \includegraphics[width=.48\linewidth]{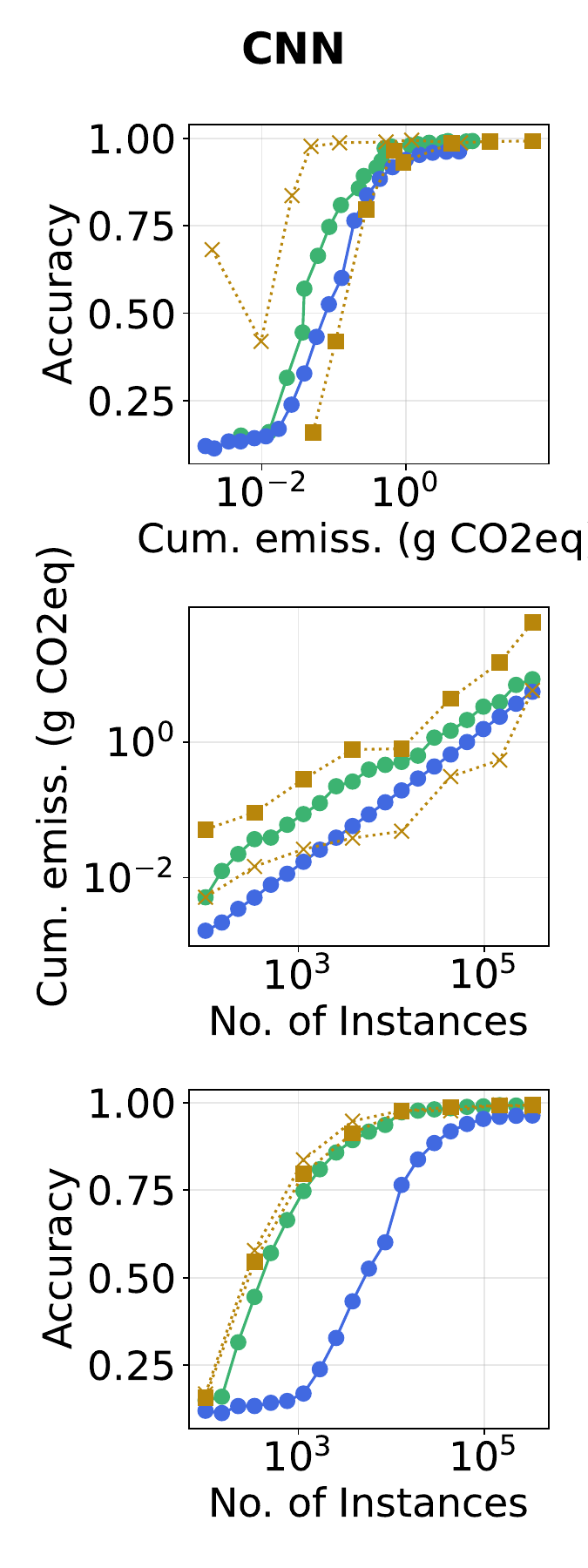}
    \includegraphics[width=.48\linewidth]{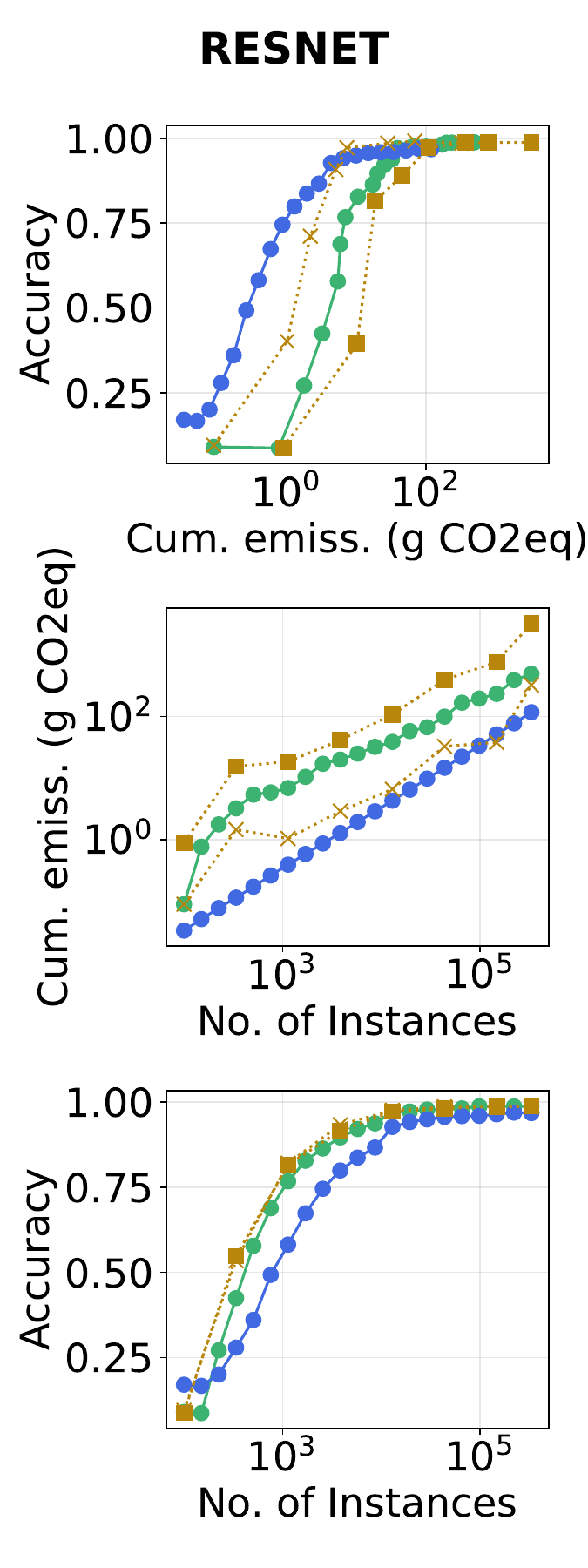}
\end{minipage}
\hfill
\begin{minipage}[c]{0.48\linewidth}
    \centering
    \textbf{ML-1M}\\
    \includegraphics[width=.48\linewidth]{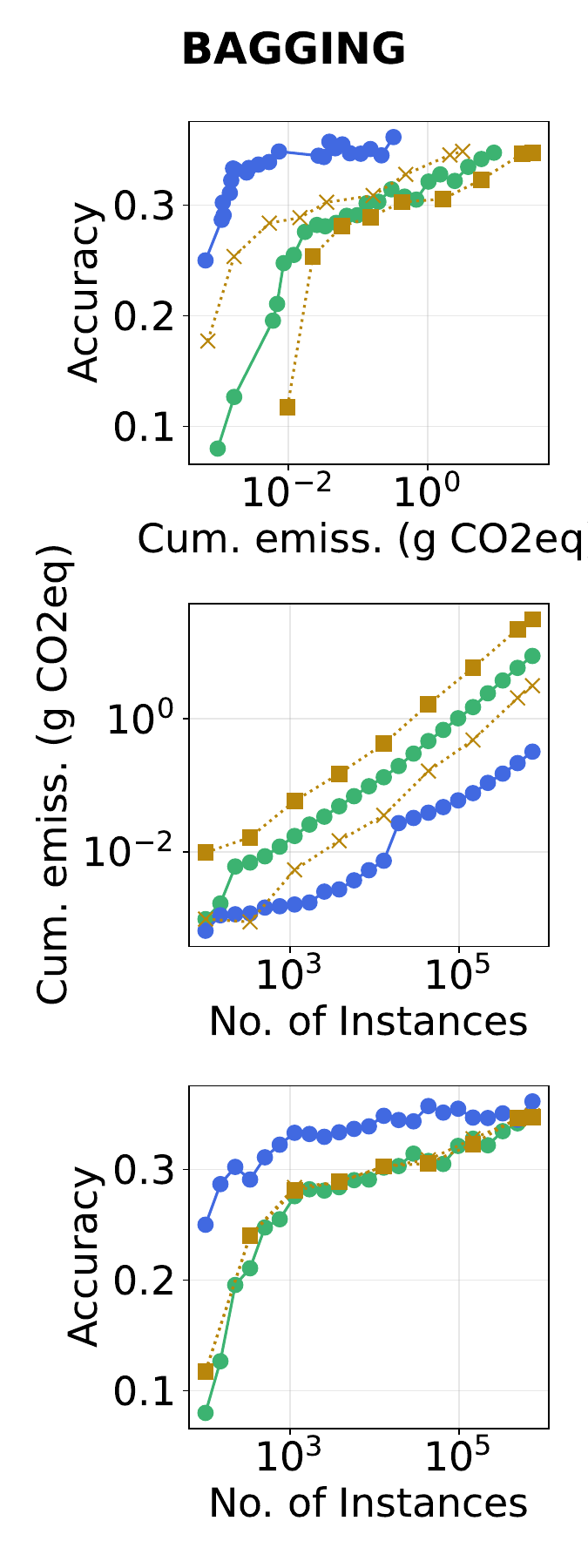}
    \includegraphics[width=.48\linewidth]{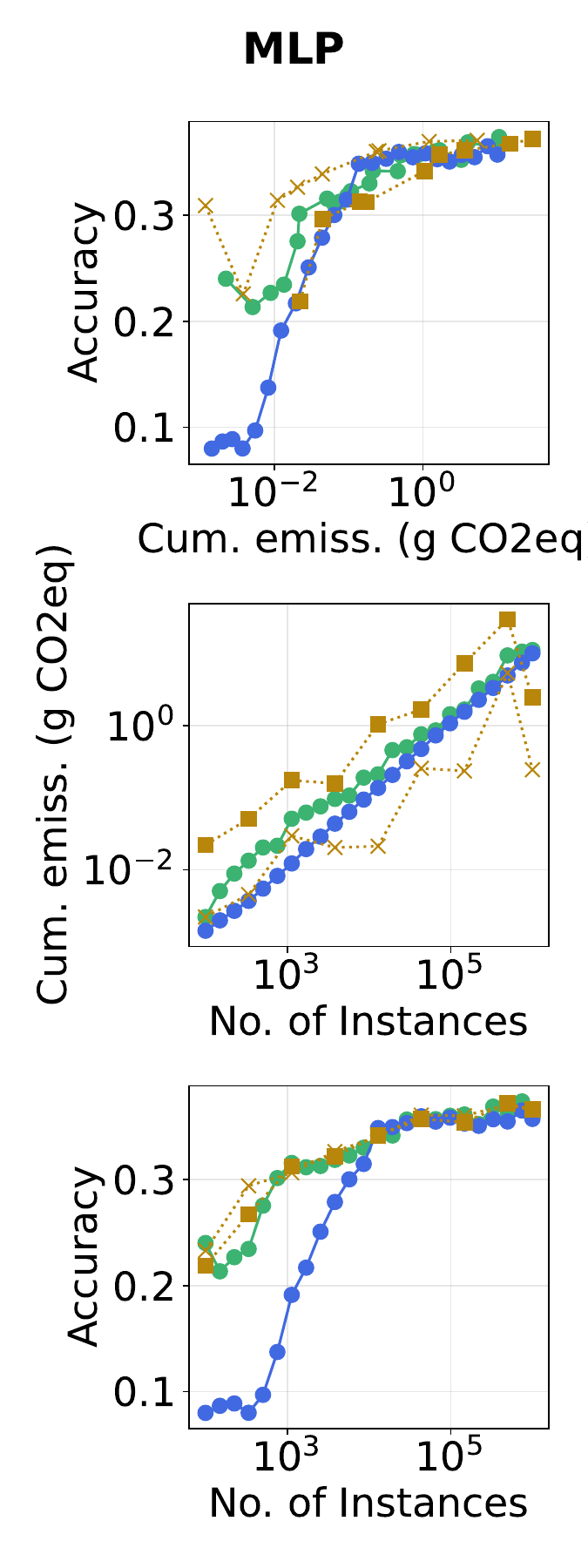}
\end{minipage}

\caption{Sustainability vs. number of instances vs. performance trade-offs for our proposal and existing naive experimental approaches on the QMNIST and ML-1M 40 datasets.}
\label{fig:results_QMNIST_ML1M}
\end{figure}

\begin{figure}[!h]
    \centering
    \textbf{KDDCUP'99 \\}
    \includegraphics[width=.30\linewidth]{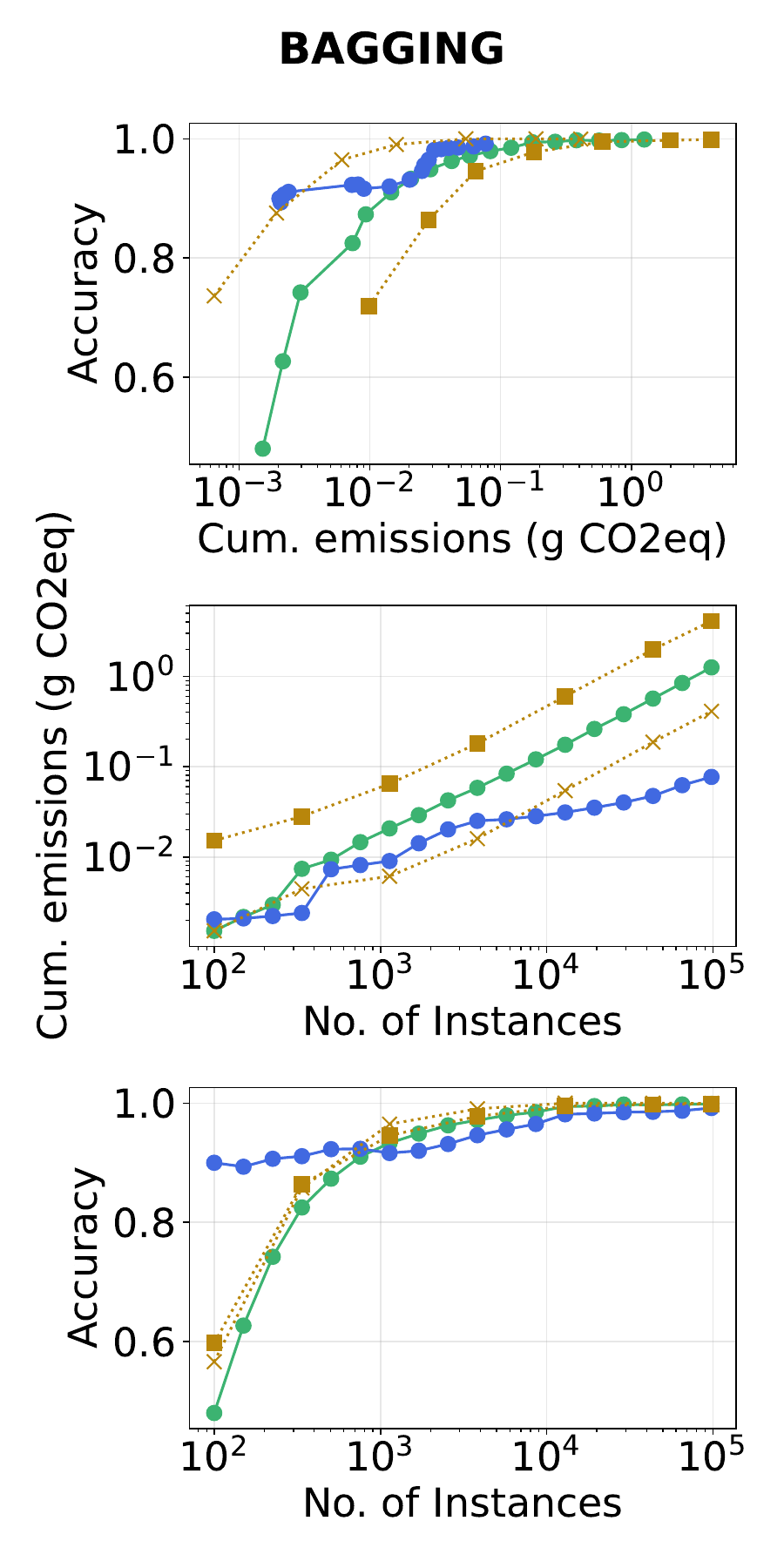}
    \includegraphics[width=.30\linewidth]{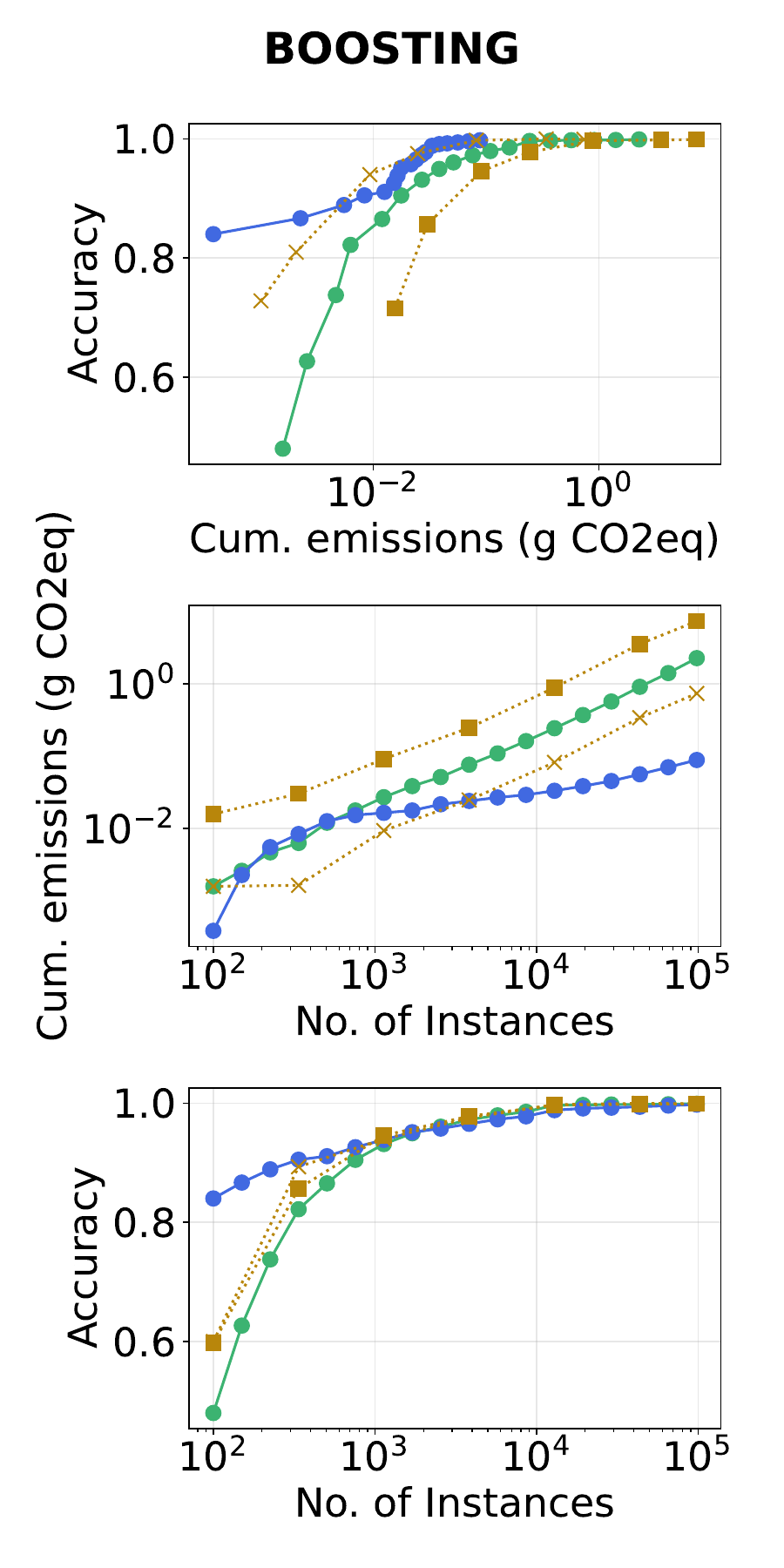}
    \includegraphics[width=.30\linewidth]{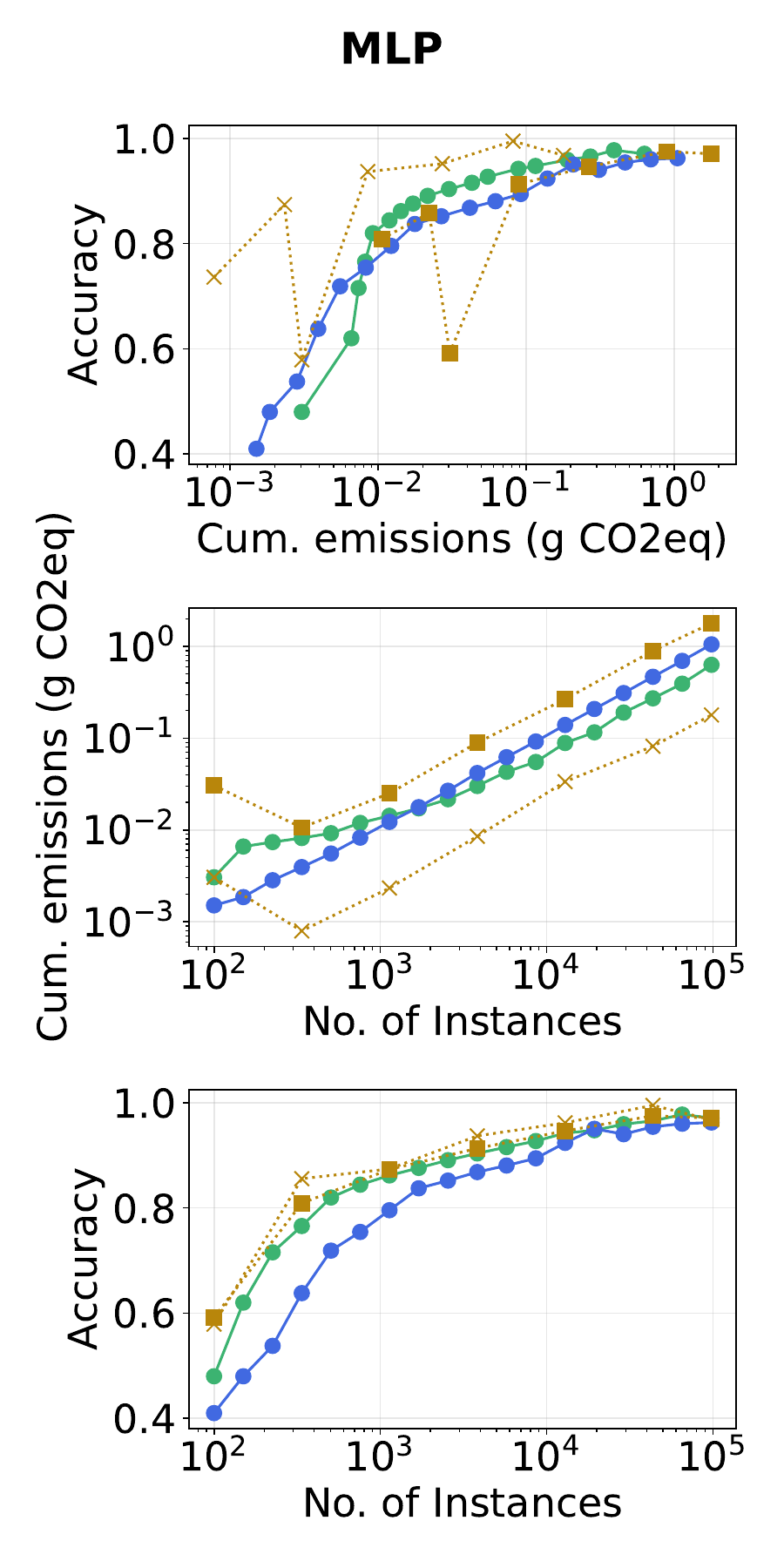}
    \caption{Sustainability vs. No. of instances vs. Performance trade-offs for our proposal and existing naive experimental approaches on the KDDCUP dataset.}
    \label{fig:results_kddcup}
\end{figure}

\begin{figure}[!h]
    \centering
    \textbf{Year Prediction MSD\\}
    \includegraphics[width=.30\linewidth]{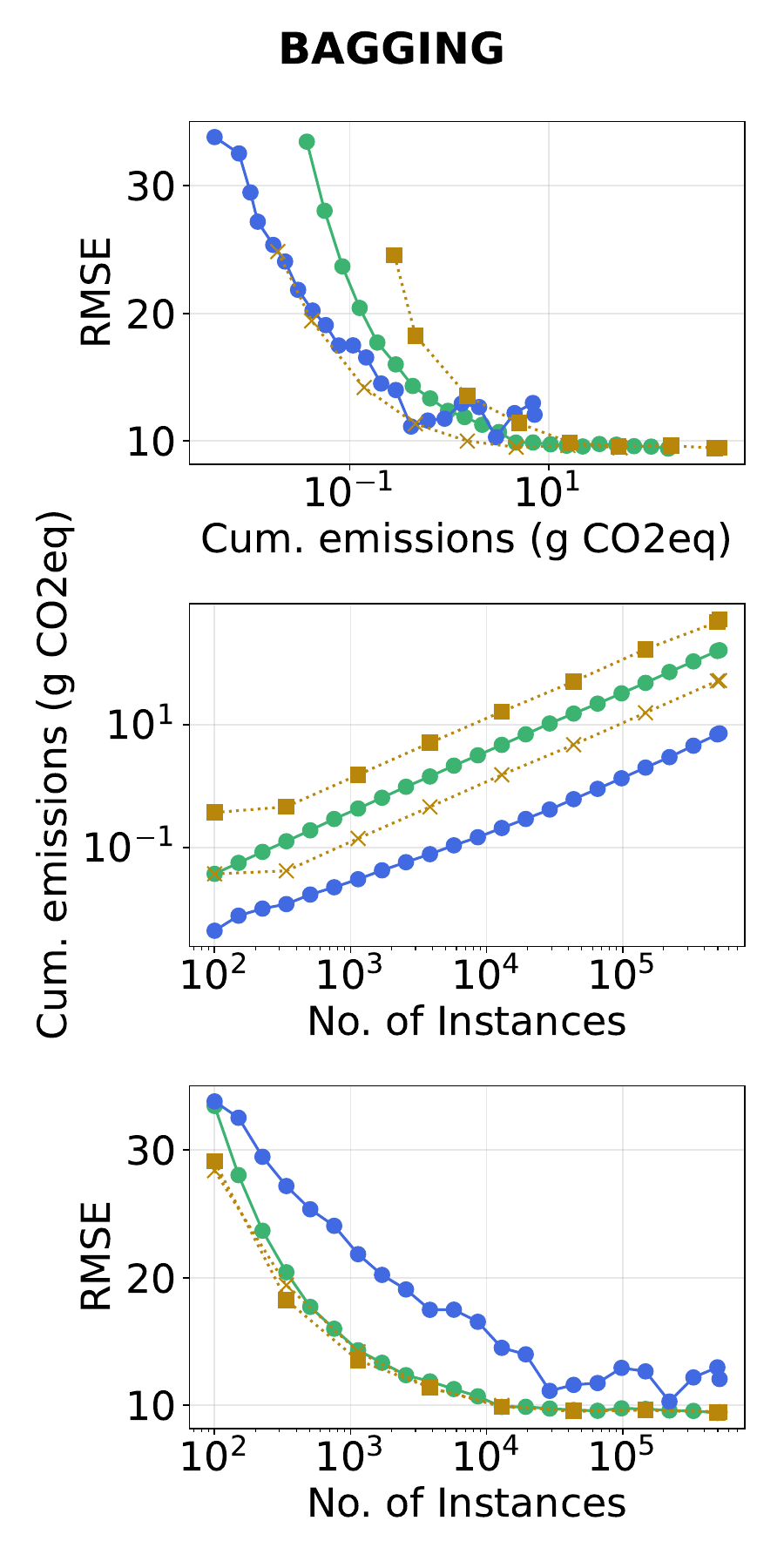}
    \includegraphics[width=.30\linewidth]{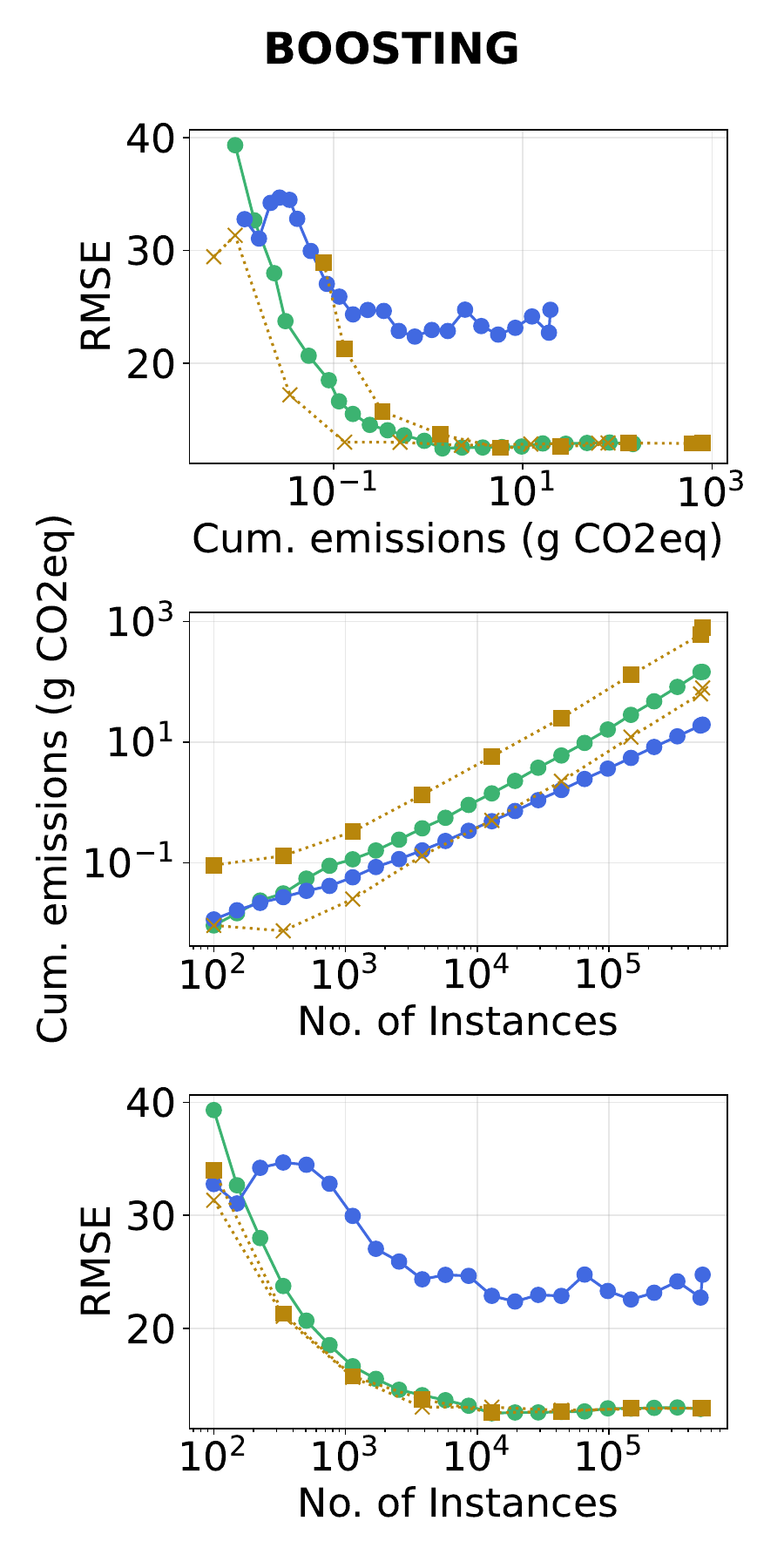}
    \includegraphics[width=.30\linewidth]{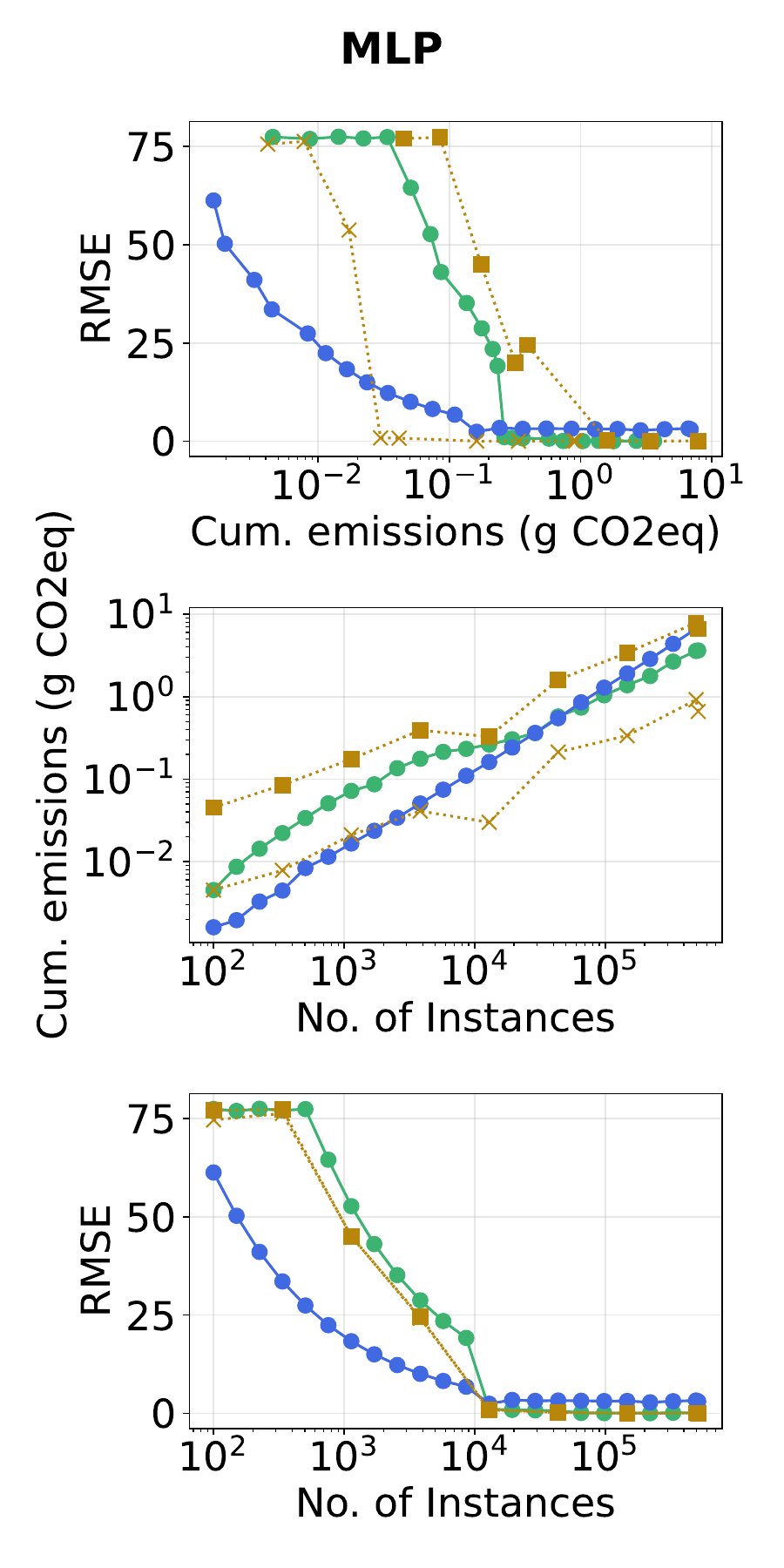}
    \caption{Sustainability vs. No. of instances vs. Performance trade-offs for our proposal and existing naive experimental approaches on the Waveform 40 dataset.}
    \label{fig:results_ypmsd}
\end{figure}

\begin{figure}[!h]
    \centering
    \textbf{CARS\\}
    \includegraphics[width=.30\linewidth]{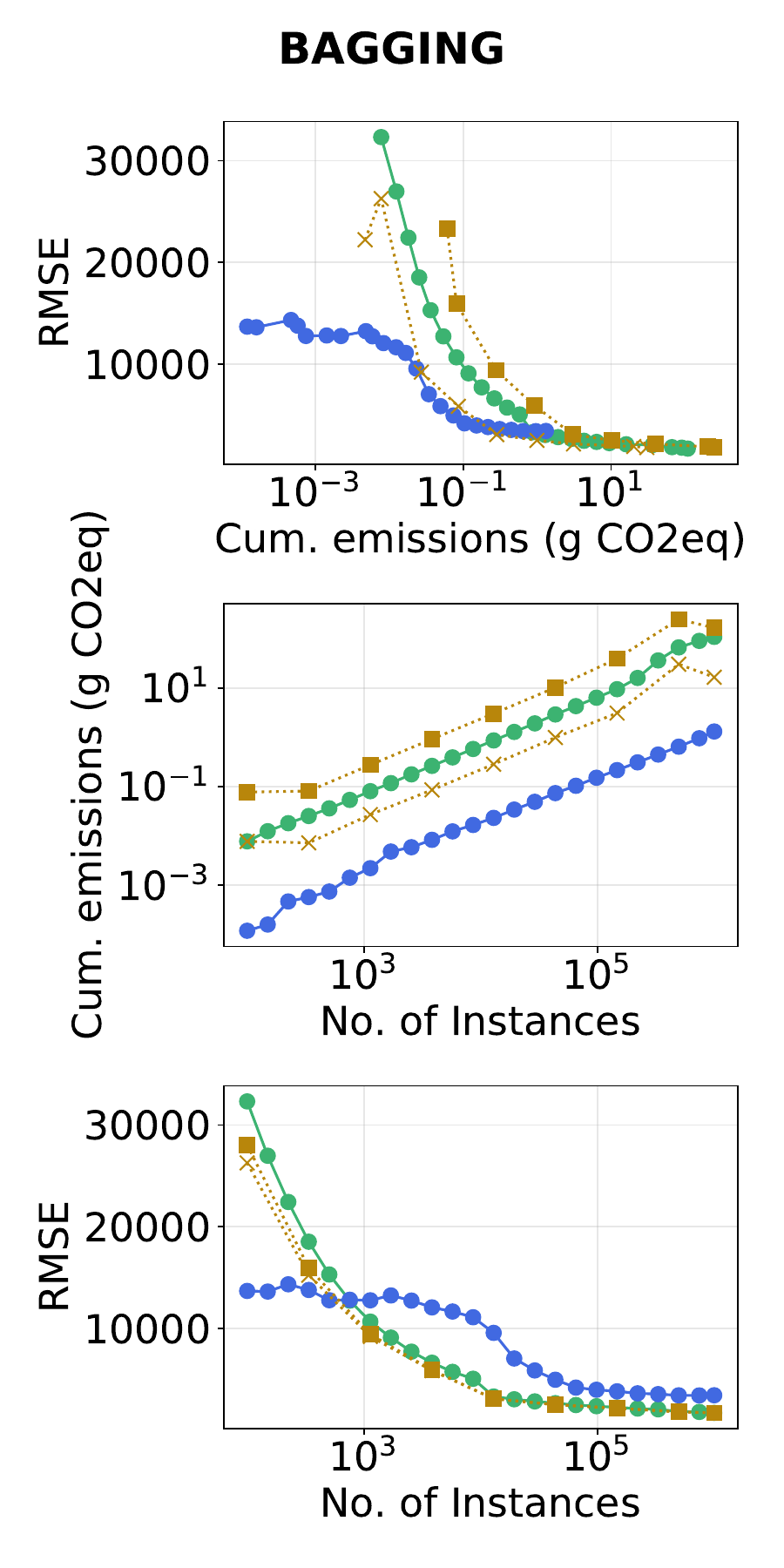}
    \includegraphics[width=.30\linewidth]{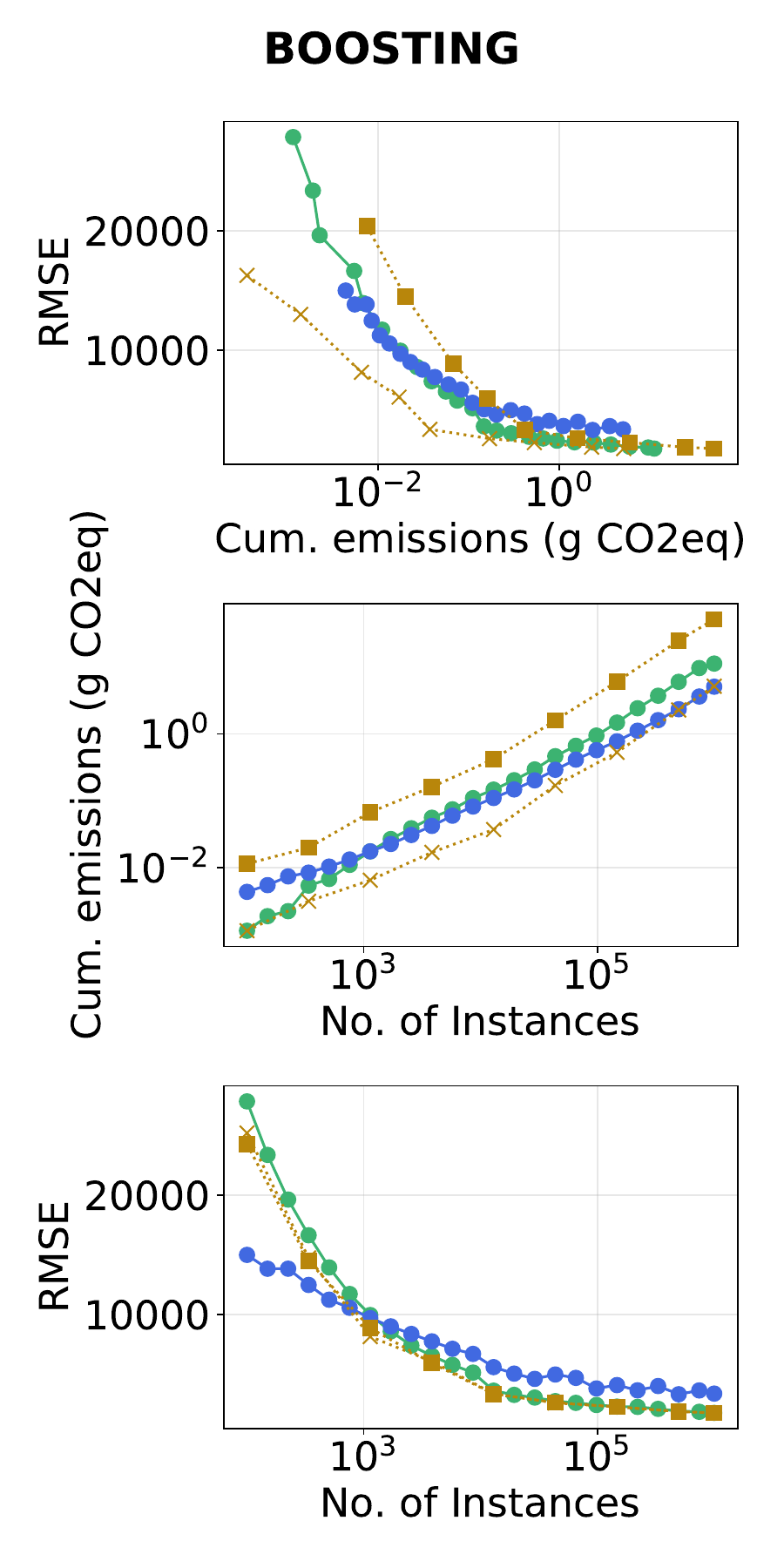}
    \includegraphics[width=.30\linewidth]{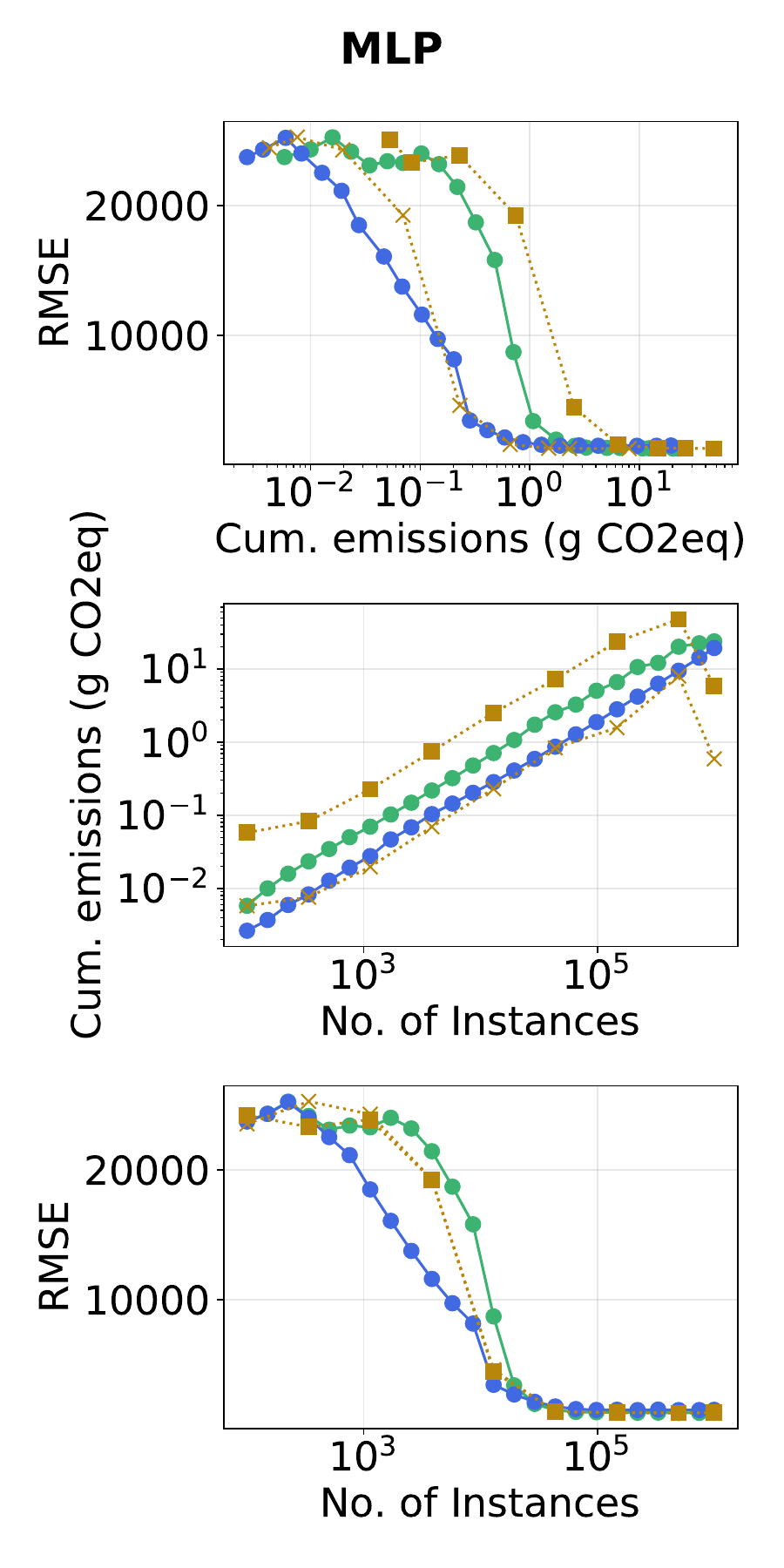}
    \caption{Sustainability vs. No. of instances vs. Performance trade-offs for our proposal and existing naive experimental approaches on the Automotive Price Prediction dataset.}
    \label{fig:results_cars}
\end{figure}

\begin{figure}[!h]
    \centering
    \textbf{ML-1M (Ranking)\\}
    \includegraphics[width=.23\linewidth]{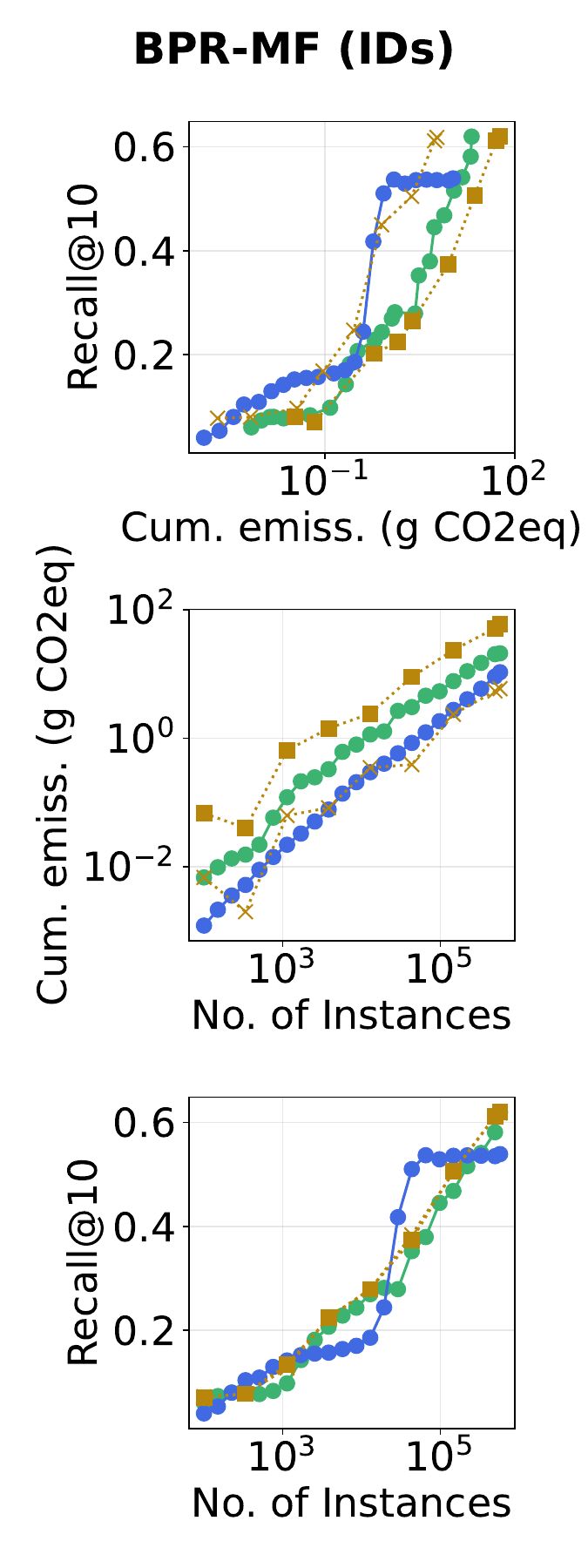}
    \includegraphics[width=.23\linewidth]{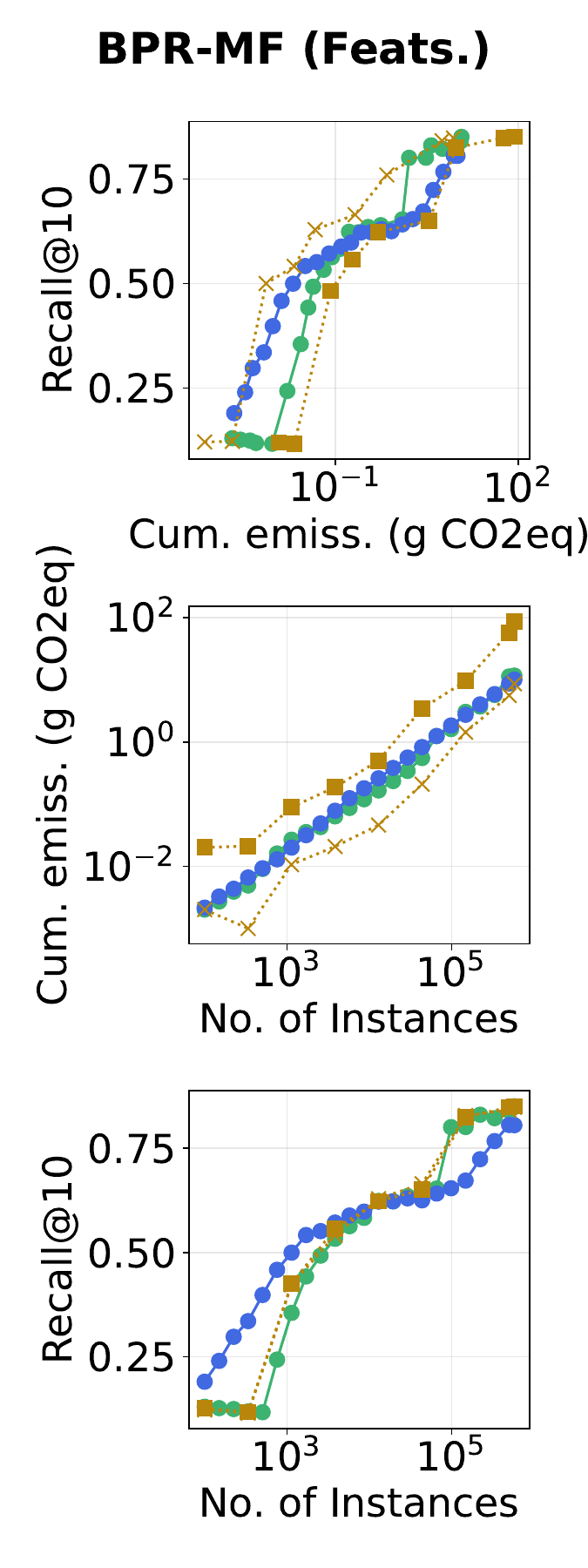}
    \includegraphics[width=.23\linewidth]{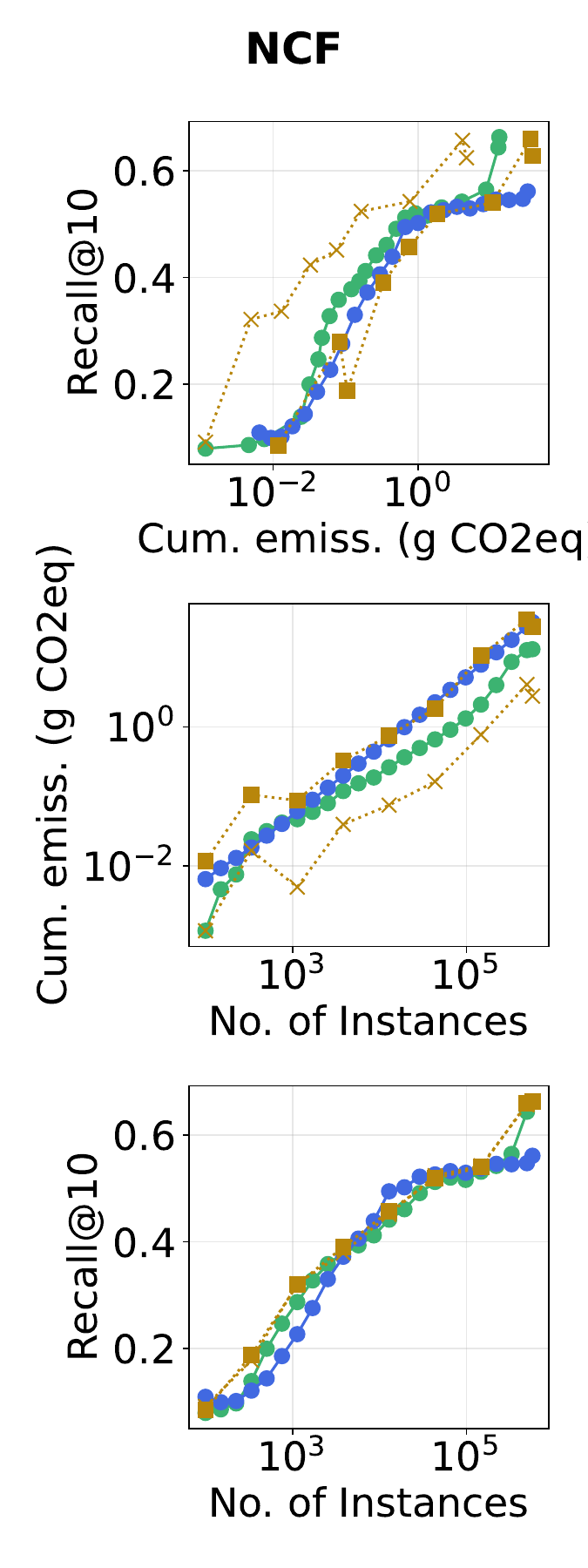}
    \includegraphics[width=.23\linewidth]{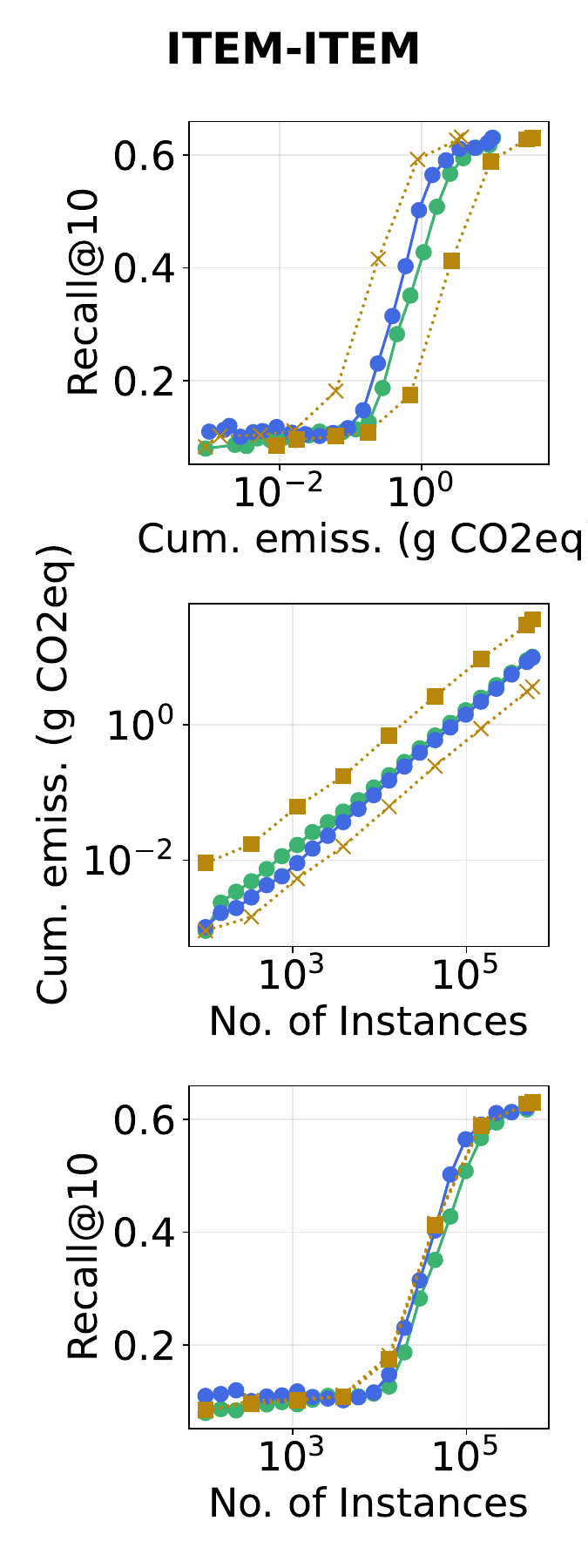}
    \caption{Sustainability vs. No. of instances vs. Performance trade-offs for our proposal and existing naive experimental approaches on the ML-1M (Ranking) dataset.}
    \label{fig:results_ml1mrank}
\end{figure}

\begin{figure}[!h]
    \centering
    \textbf{NETFLIX\\}
    \includegraphics[width=.30\linewidth]{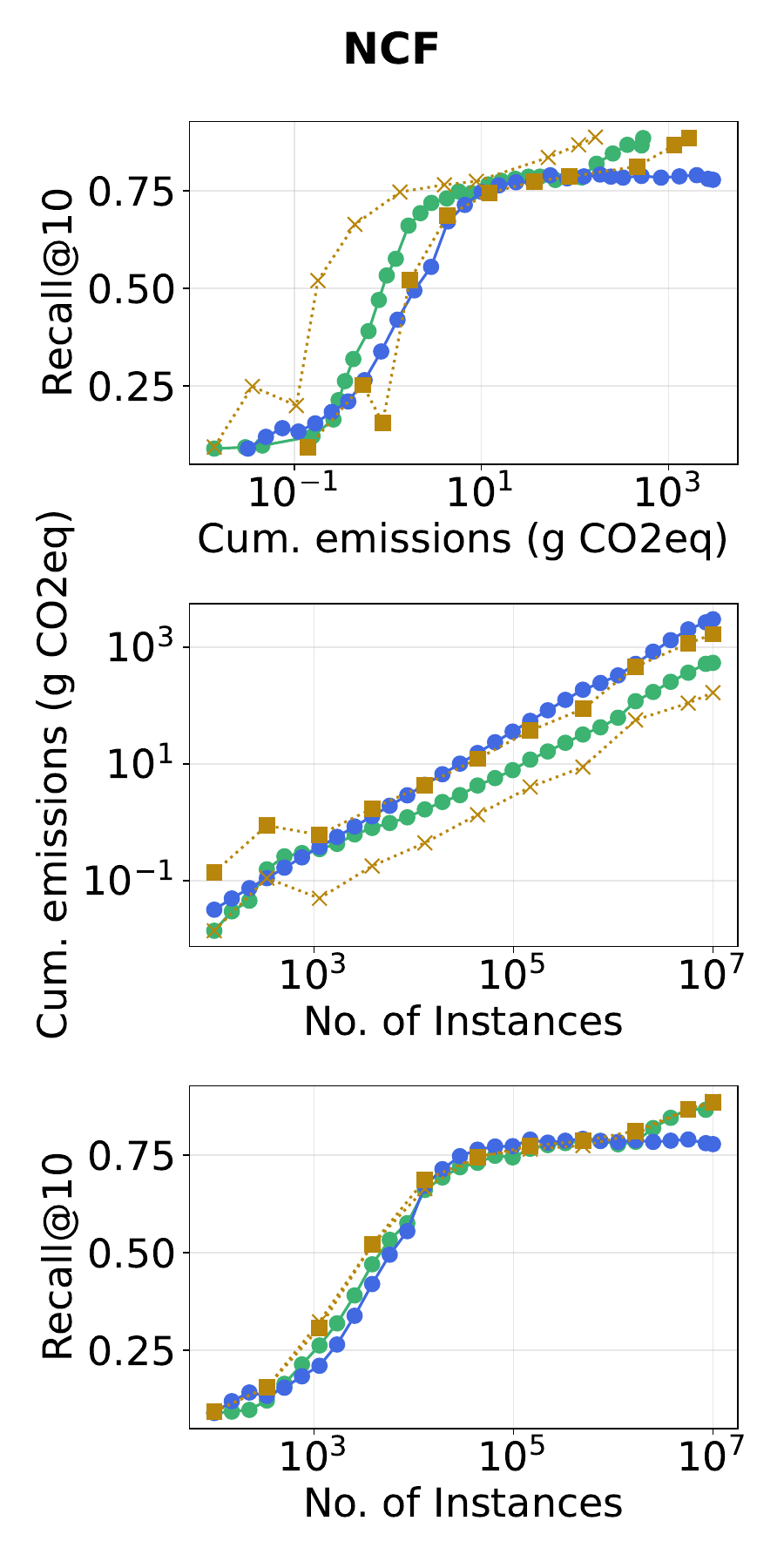}
    \includegraphics[width=.30\linewidth]{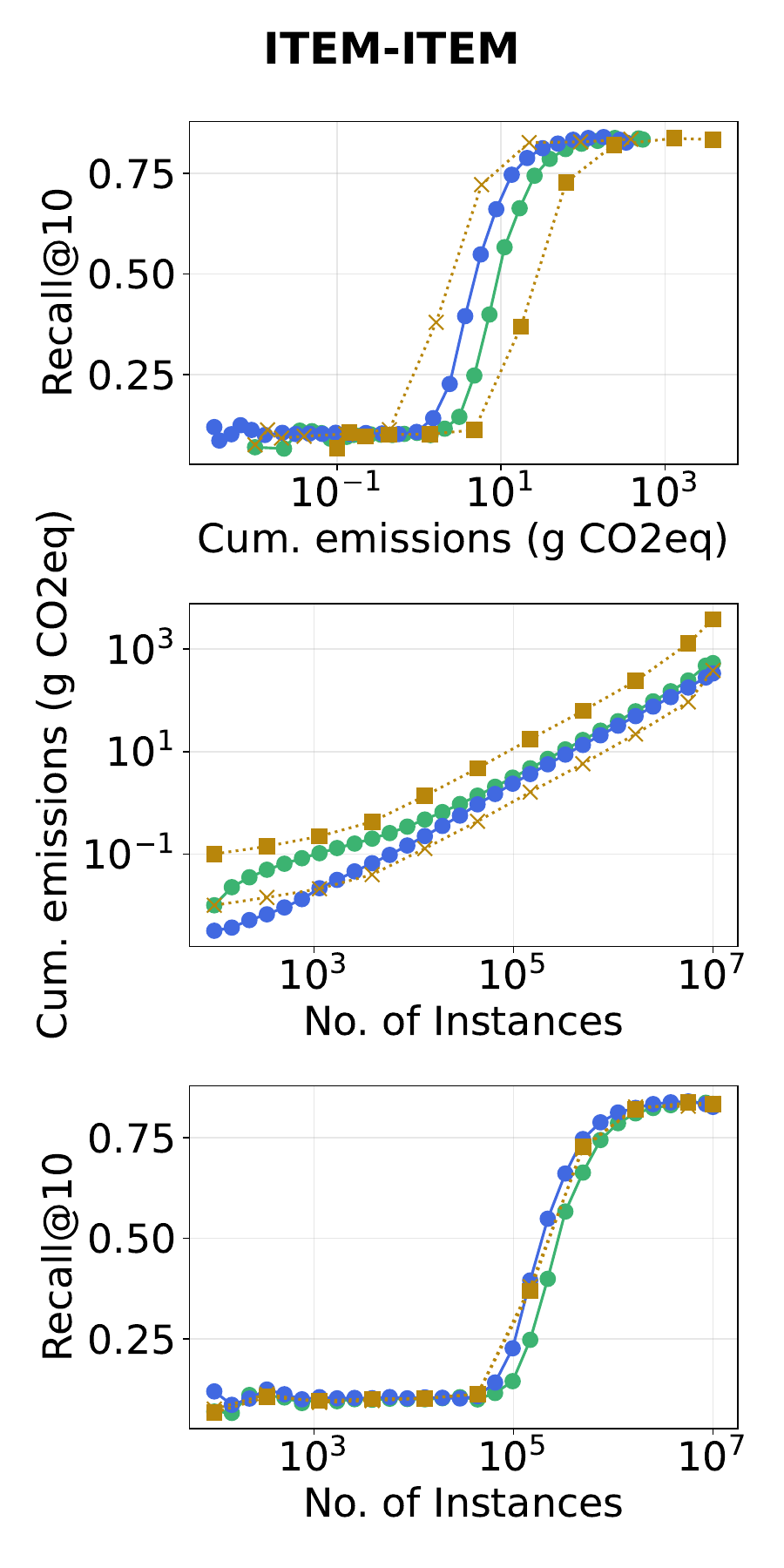}
    \includegraphics[width=.30\linewidth]{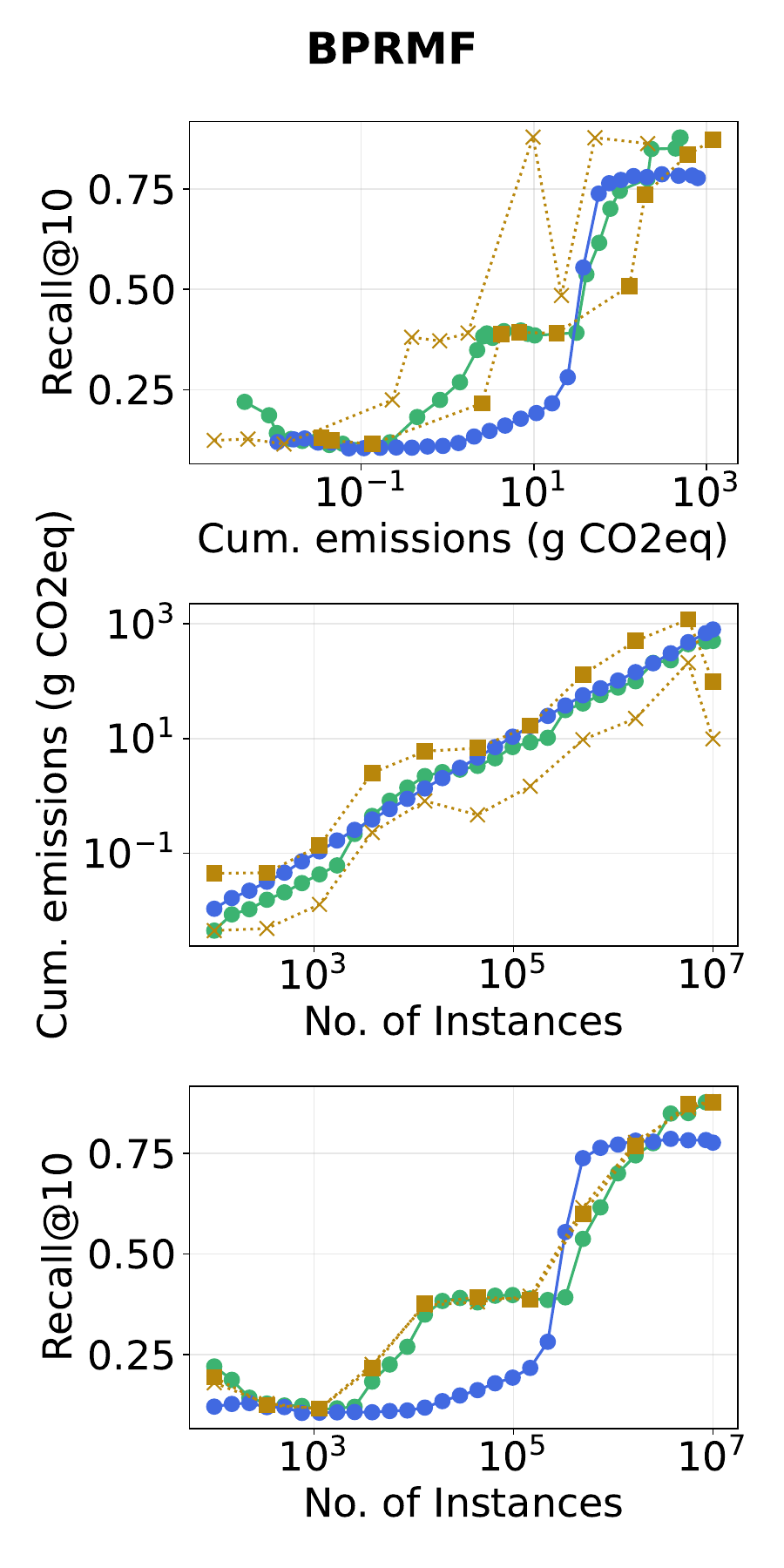}
    \caption{Sustainability vs. No. of instances vs. Performance trade-offs for our proposal and existing naive experimental approaches on the Netflix dataset.}
    \label{fig:results_netflix}
\end{figure}

\begin{figure}[!h]
    \centering
    \textbf{CELEB-A\\}
    \includegraphics[width=.30\linewidth]{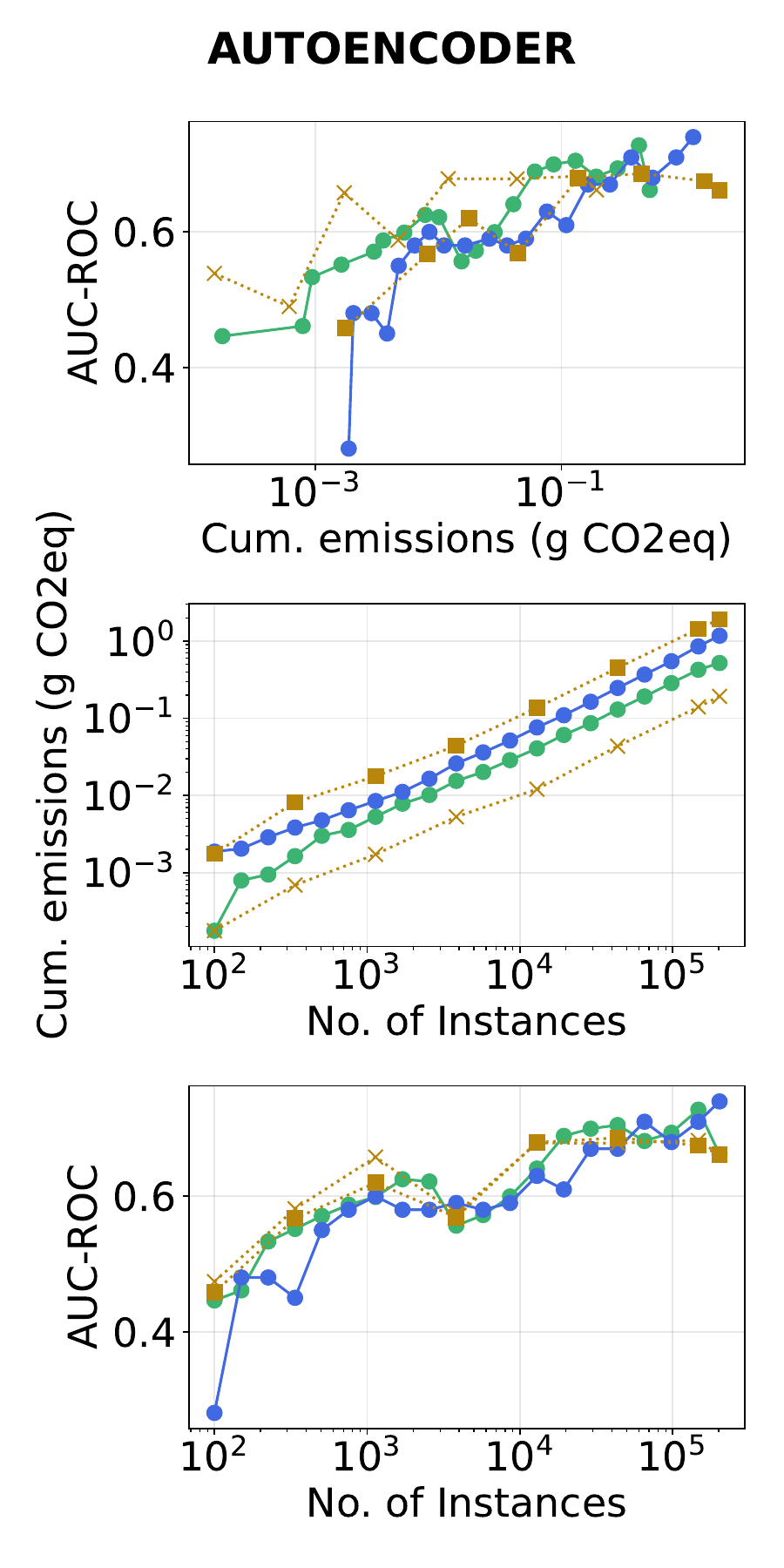}
    \includegraphics[width=.30\linewidth]{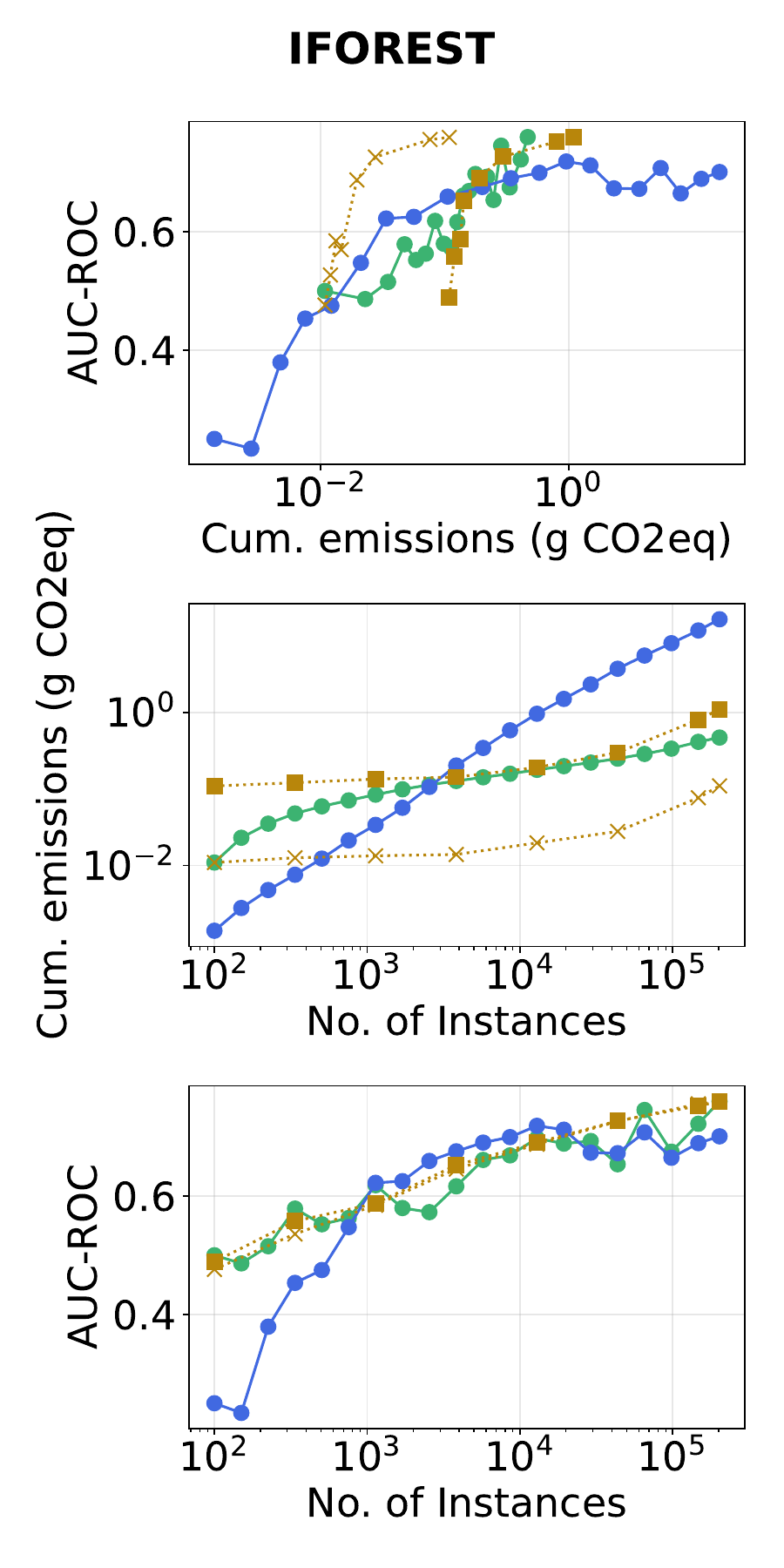}
    \includegraphics[width=.30\linewidth]{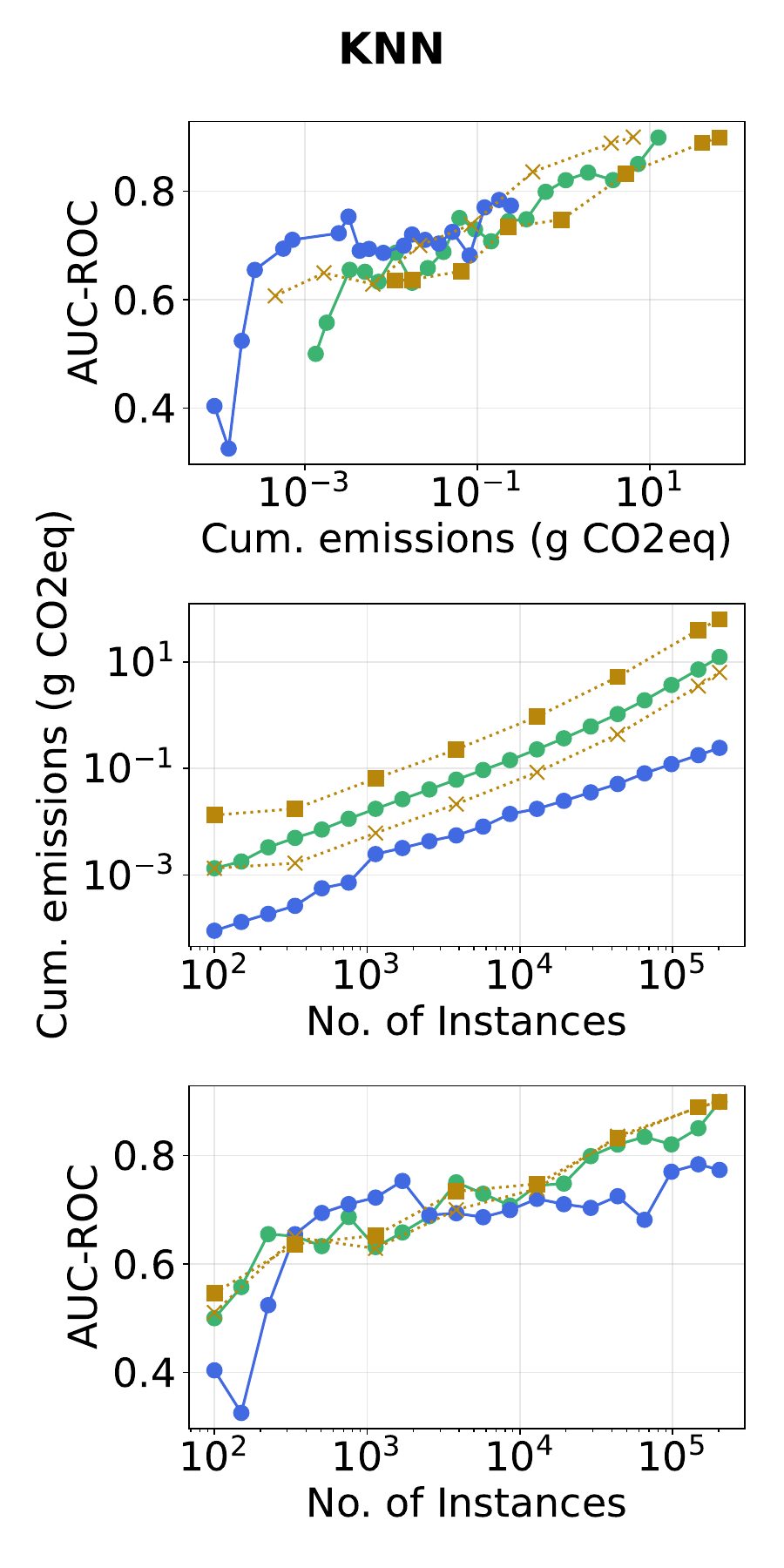}
    \caption{Sustainability vs. No. of instances vs. Performance trade-offs for our proposal and existing naive experimental approaches on the CelebA dataset.}
    \label{fig:results_celebA}
\end{figure}

Figures \ref{fig:results_waveform}-\ref{fig:results_celebA} show the sustainability vs. performance trade-off and the impact of the number of instances on the performance and sustainability of batch and streaming models under our proposed long-term sustainability evaluation protocols. 

Concerning these performance vs. sustainability results, it is worth noting that our objective is not to declare a victor in the Streaming vs. Batch comparison, nor to establish a best model for each dataset or task; instead, we seek to demonstrate the importance of a robust sustainability analysis by incorporating the more strict requirements of Streaming tasks into a more realistic sustainability evaluation that is representative of the long-term use that is desirable for an ML pipeline.

The results of our computational experiments reveal several interesting insights that prove initial assumptions about the current landscape of Green AI evaluation:

\begin{itemize}
    \item We observe that performing non-cumulative, sterile measurements of sustainability in individual Holdout or Cross-Validation experiments will severely underestimate or overestimate (respectively) the real cumulative, long-term environmental impact of maintaining a model. This is in line with our starting argument that such sustainability evaluation protocols are measuring the environmental impact of performance-measurement experiments, and not the sustainability of long-term model usage itself. 
    \item In occasions, the results provided by sterile measurements on ``frozen'' holdout or CV procedures are erratic; this is more easily seen on NN-based models (e.g., in MLP models on KDDCUP, Fig. \ref{fig:results_kddcup}), typically caused by shorter or longer epoch and early stopping runs. Contrarily, our proposal appropiately reflects the incremental environmental cost of the AI life-cycle typically observed in Online models, allowing it to naturally account for this behaviour.
    \item Models based on NN's have consistently rough starts w.r.t. to the sustainability vs. performance trade-off, something that is not apparent if using traditional evaluation protocols. Our proposal is able to highlight that, when starting from a few examples in a realistic setting, decisions should be made whether to wait for a higher initial number of available instances if delaying the start of predictions is possible, or whether sustainability has to be sacrificed to be able to perform predictions from the beginning. 
\end{itemize}

Overall, our proposed ML sustainability evaluation protocol is able to more accurately capture the long-term tendencies of the environmental impact of models, unlike existing protocols and experimental practices. 
\section{Conclusions}

This work provides two key contributions in the field of Green AI and the broader scope of sustainability in Machine Learning, by 1) articulating the existing issues of short-term, narrow-minded experimental methods for evaluating sustainability trade-offs in ML models, and 2) providing a robust evaluation methodology that is model-agnostic, is representative of the longer-term ML lifecycle, and can be easily adapted for more specialized use-cases. Our protocol is based on the reasonable assumptions that in most real-world ML tasks, data is incremental and potentially evolving, and that models need to be constantly revised to leverage the knowledge from recent data. 

With regards to future work, we consider that there are three aspects to be explored that stem from our research:

\begin{itemize}
    \item Drifting environments should be considered (at present, we consider non-evolving data sources), as these are ubiquitous in everyday ML tasks for the real-world, and will have effects on the design of the retraining/reevaluation policies used by our algorithm. 

    \item Our experiments assume immediate and complete labelling of incoming examples, as well as infinite storage, which are ideal conditions. While accounting these factors would not hinder the applicability of our method, they would affect the precise comparison of different models when employing our proposed evaluation protocol.

    \item Our experiments' dataset and model choices on task and model diversity, prioritizing simple models and strong baselines that have known properties and established batch and streaming variants; while this is best for proving the properties of our algorithm, we theorize that the impact of our method's observations on real-life massive-scale data and models, such as LLMs, will be even more notable.  
\end{itemize}

\begin{credits}
\subsubsection{\ackname} This research work has been funded by MICIU/AEI/ 10.13039/ 501100011033 and ESF+ (FPU21/05783), ERDF A way of making Europe (PID2019-109238GB-C22, PID2023-147404OB-I00), and by the Xunta de Galicia (ED431C 2022/44) with the European Union ERDF funds. CITIC, as Research Center accredited by Galician University System, is funded by ``Consellería de Cultura, Educación e Universidade from Xunta de Galicia", supported in an 80\% through ERDF Operational Programme Galicia 2021-2027, and the remaining 20\% by ``Secretaría Xeral de Universidades" (ED431G 2023/01)

\subsubsection{\discintname}
The authors have no competing interests to declare that are
relevant to the content of this article.
\end{credits}
%
%
\bibliographystyle{splncs04}
\bibliography{main}

\end{document}